\documentclass[conference]{IEEEtran}
\usepackage{times}

\usepackage[numbers]{natbib}
\usepackage{multicol}
\usepackage[bookmarks=true]{hyperref}
\usepackage{subfiles}
\usepackage{amssymb}
\usepackage{amsmath}
\usepackage{bm}
\usepackage[table]{xcolor}
\usepackage[linesnumbered,lined,ruled]{algorithm2e}
\usepackage{graphicx}
\usepackage{booktabs} 
\usepackage{tablefootnote}
\usepackage{enumerate}
\usepackage{balance}
\usepackage{comment}

\makeatletter
\newcommand{\nosemic}{\renewcommand{\@endalgocfline}{\relax}}
\newcommand{\dosemic}{\renewcommand{\@endalgocfline}{\algocf@endline}}
\newcommand{\pushline}{\Indp}
\newcommand{\popline}{\Indm\dosemic}
\makeatother

\newcommand{\subjto}{\mathrm{s.t.}}

\newcommand{\bmtau}{\bm{\tau}}
\newcommand{\bmv}{\bm{v}}
\newcommand{\bmq}{\bm{q}}

\newcommand{\calP}{\mathcal{P}}
\newcommand{\calQ}{\mathcal{Q}}

\newcommand{\calL}{\mathcal{L}}

\newcommand{\calK}{\mathcal{K}}

\newcommand{\RR}{\mathbb{R}}

\DeclareMathOperator*{\argmin}{arg\,min}

\definecolor{pastelblue}{rgb}{0.68, 0.78, 0.81}

\begin{document}

\title{From Compliant to Rigid Contact Simulation: \\ a Unified and Efficient Approach}


\author{\authorblockN{Justin Carpentier\textsuperscript{*}}
\authorblockA{
Inria, \'Ecole normale supérieure \\ CNRS, PSL Research University\\
75005 Paris, France\\
Email: \href{mailto:justin.carpentier@inria.fr}{justin.carpentier@inria.fr}}
\and
\authorblockN{Quentin {Le Lidec}\textsuperscript{*}}
\authorblockA{
Inria, \'Ecole normale supérieure \\ CNRS, PSL Research University\\
75005 Paris, France\\
Email: \href{mailto:quentin.le-lidec@inria.fr}{quentin.le-lidec@inria.fr}}
\and
\authorblockN{Louis Montaut\textsuperscript{*}}
\authorblockA{
Inria, \'Ecole normale supérieure \\ CNRS, PSL Research University\\
75005 Paris, France\\
Email: \href{mailto:louis.montaut@inria.fr}{louis.montaut@inria.fr}}
}

\maketitle
\def\thefootnote{*}\footnotetext{Equal contribution. Authors are listed in alphabetical order.}\def\thefootnote{\arabic{footnote}}

\begin{abstract}

Whether rigid or compliant, contact interactions are inherent to robot motions, enabling them to move or manipulate things. 
Contact interactions result from complex physical phenomena, that can be mathematically cast as Nonlinear Complementarity Problems (NCPs) in the context of rigid or compliant point contact interactions.
Such a class of complementarity problems is, in general, difficult to solve both from an optimization and numerical perspective. 
Over the past decades, dedicated and specialized contact solvers, implemented in modern robotics simulators (e.g., Bullet, Drake, MuJoCo, DART, Raisim) have emerged.
Yet, most of these solvers tend either to solve a relaxed formulation of the original contact problems (at the price of physical inconsistencies) or to scale poorly with the problem dimension or its numerical conditioning (e.g., a robotic hand manipulating a paper sheet).
In this paper, we introduce a unified and efficient approach to solving NCPs in the context of contact simulation.
It relies on a sound combination of the Alternating Direction Method of Multipliers (ADMM) and proximal algorithms to account for both compliant and rigid contact interfaces in a unified way.
To handle ill-conditioned problems and accelerate the convergence rate, we also propose an efficient update strategy to adapt the ADMM hyperparameters automatically.
By leveraging proximal methods, we also propose new algorithmic solutions to efficiently evaluate the inverse dynamics involving rigid and compliant contact interactions, extending the approach developed in MuJoCo.  
We validate the efficiency and robustness of our contact solver against several alternative contact methods of the literature and benchmark them on various robotics and granular mechanics scenarios.
Overall, the proposed approach is shown to be competitive against classic methods for simple contact problems and outperforms existing solutions on more complex scenarios, involving tens of contacts and poor conditioning.
Our code is made open-source at \url{https://github.com/Simple-Robotics/Simple}.
\end{abstract}

\IEEEpeerreviewmaketitle

\section{Introduction}
\label{sec:introduction}

Contact interactions are the substrate of movement.
When performing locomotion or manipulation tasks, one naturally makes and breaks contact with the environment.
Providing robots with such abilities would require them to precisely apprehend the physics of contact.
In this respect, simulators have been key components for various robotics applications.
Reinforcement Learning (RL) policies for manipulation \cite{chen2022system,handa2023dextreme} or locomotion \cite{lee2020learning} often rely on years of simulated trajectories during training.
Alternatively, Model Predictive Control (MPC) techniques call the simulator and its derivatives at high frequency at runtime to achieve a reactive behavior \cite{koenemann2015whole,jenelten2024dtc}.

Simulating rigid contact interactions with friction requires solving a Nonlinear Complementarity Problem (NCP) which has been recognized as a hard problem both from an optimization and numerical perspective~\cite {acary2017contact}.
For this reason, simulators proceed to tradeoffs between physical realism, robustness, and efficiency.
In this respect, each physics engine embeds its own contact model and contact solver, which come with their underpinning physical hypotheses and numerical capabilities.

Earlier simulators like ODE~\cite{ode:2008} and Bullet~\cite{coumans2021} have been developed for graphical purposes. 
Graphical applications mainly require the physics engine to provide visually consistent trajectories.
Nowadays, the requirements in robotics applications are different: the physics of contacts should be as close as possible to reality, and the contact solver should be numerically efficient.
Indeed, this would reduce training time and enhance the transferability of RL while making MPC more reactive.

\begin{figure}[!t]
    \centering
    \vspace{-1cm}
    \includegraphics[width=.95\linewidth]{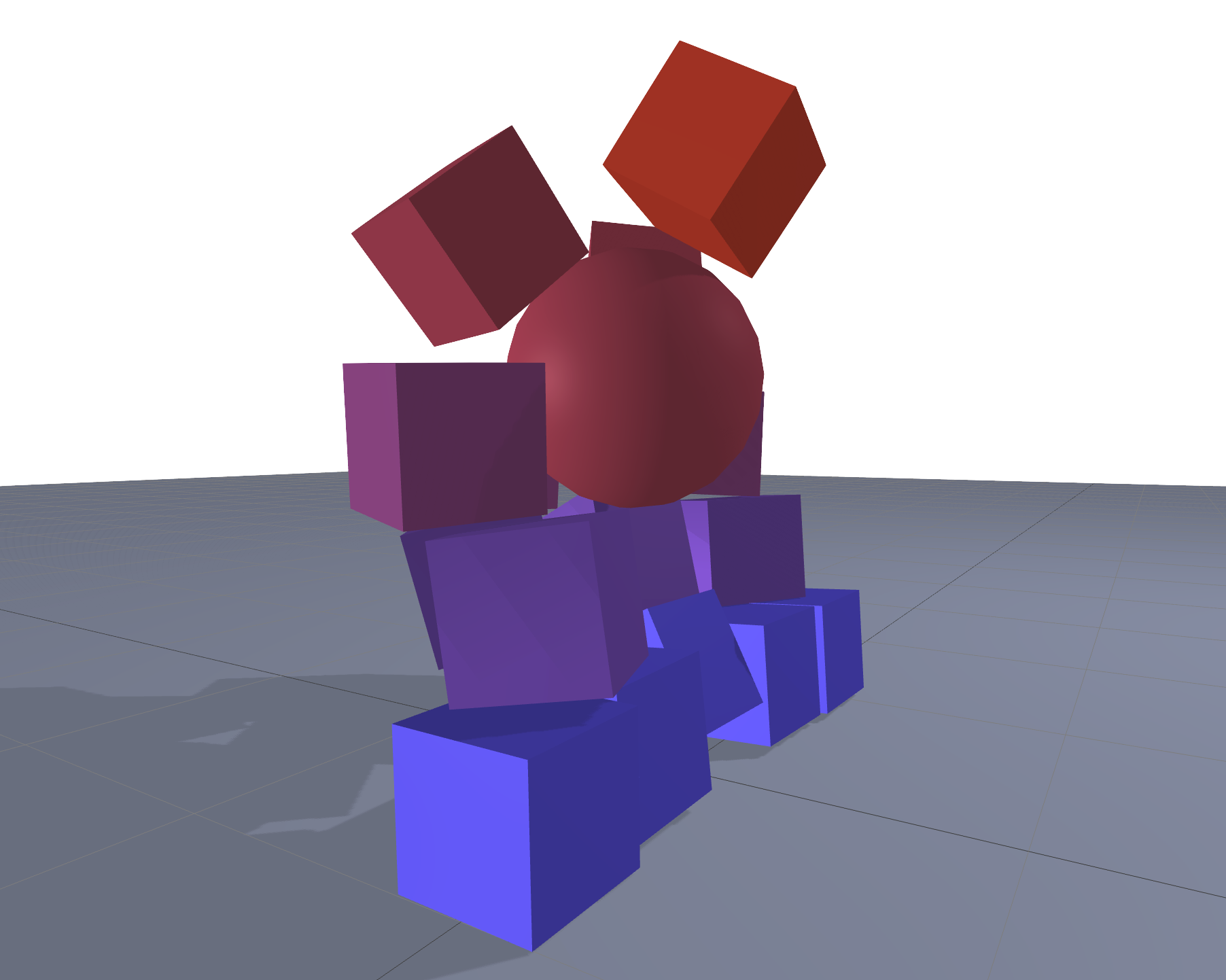}
    \caption{\textbf{Simulation of an ill-conditioned stack of cubes.} 
    Our approach robustly solves high-dimensional and ill-conditioned problems such as the ones involved when simulating a stack of $15$ cubes involving more than $60$ contact points and objects' masses varying from $1$\,kg (bottom row, in light blue) to $10.000$\,kg (highest row, in bright red). 
    Simulation videos at \textbf{\url{https://youtu.be/i_qg9cTx0NY?si=NGtx1tiYrIGtHXSK}}.
    }
    \label{fig:header}
    \vspace{-1em}
\end{figure}

More recently, robotics-oriented engines were developed with a focus on efficiency.
MuJoCo \cite{todorov2012mujoco} introduces a physical relaxation that induces a convex contact model with distant and compliant interactions.
This choice enables the use of powerful algorithms from the optimization literature and leads to improved stability of the simulation.
Its robustness has made MuJoCo the standard testbed for RL algorithms, but its reality gap still limits its use in robotics.
Drake \cite{drake} leverages a similar contact model \cite{castro2022unconstrained} and proposes a procedure to set the compliance parameter more realistically, according to the physical parameters of the simulated objects.
This results in an improved sim-to-real transfer capability on manipulation tasks \cite{TRI2023video}.
RaiSim \cite{raisim} fixes MuJoCo's artifacts due to distant and artificially compliant contact forces but at the expense of less robust numerics~\cite{lelidec2023contact}.
The resulting contact model drove success in quadrupedal locomotion in challenging setups \cite{hwangbo2019learning,lee2020learning,miki2022learning}.

Recent works \cite{horak2019similarities,lelidec2023contact} compare the different existing approaches and their impact.
Although they highlight the significant improvements in contact simulation driven by the efforts of the robotics community, they also reveal current simulators are all subject to trade-offs and, thus, some barriers are still to be removed.
Some experimental works~\cite{fazeli2017empirical,acosta2022validating} also evaluate contact models from current simulators against real-world data.
They show that they still fail to capture accurately the physical phenomena involved in contact interactions.
Inspired by these studies, we aim to improve the physicality of current simulators while preserving their efficiency.

\noindent We make the following contributions:
\begin{itemize}
    \item we introduce a new formulation and related algorithm for solving forward dynamics problems with contacts that can efficiently deal with compliant and rigid contacts without making any physical relaxation,
    \item we propose a formulation and an algorithmic solution for computing inverse dynamics with rigid and compliant contacts, the reciprocal of forward dynamics, extending the seminal work of Todorov~\cite{todorov2012mujoco},
    \item we provide an open-source and efficient C++ implementation leveraging the Eigen library~\cite{eigenweb}, and that can be easily plugged into existing simulation frameworks,
    \item we extensively evaluate the proposed solutions on both mechanics and robotics scenarios of the literature.
\end{itemize}

\vspace{1em}
The paper is organized as follows. 
We first provide the necessary background on constrained dynamics, how it connects to proximal optimization, the NCP appearing during the simulation of frictional contacts and the existing solver from the robotics community (Sec.~\ref{sec:background}).
Second, we propose an ADMM-based algorithm for contact solving (Sec.~\ref{sec:admm}).
Third, we introduce a similar algorithm for the problem of inverse dynamics in the presence of contacts (Sec.~\ref{sec:inverse}). 
Finally, we evaluate our approach against standard problems of the robotics and mechanics communities (Sec.~\ref{sec:results}) before discussing how it compares to previous works (Sec.~\ref{sec:discussion}).

\section{Background} 
\label{sec:background}
In this section, we recall the principles underlying the simulation of systems under rigid and compliant constraints.
To prepare the ground for the proposed algorithm, we introduce proximal operators and how they can be interpreted from a mechanics point of view.
Finally, we introduce the NCP underlying the simulation of physical systems in the presence of frictional contacts and the existing approaches to solve it.

\subsection{Equality-constrained dynamics}

The free motion of a poly-articulated system is governed by the so-called Lagrangian dynamics of the form
\begin{equation}
    \label{eq:motion}
    M(\bm{q})\bm{\dot{v}} +b(\bm{q},\bm{v}) = \bm{\tau}
\end{equation}
where we denote by $\bm{q} \in \mathcal{Q} \cong \mathbb{R}^{n_q}$, $\bm{v} \in \mathcal{T}_q\mathcal{Q} \cong \mathbb{R}^{n_v}$ and $\bm{\tau} \in \mathcal{T}_q^*\mathcal{Q} \cong \mathbb{R}^{n_v}$, the joint configuration vector, the joint velocity
vector and the joint torque vector.
$M(\bm{q})$ is the joint-space inertia matrix and $b(\bm{q}, \bm{v})$ includes terms related to the gravity, Coriolis, and centrifugal effects.
Considering constraints such as kinematic loop closures or anchor points can be done by adding implicit equations of the form
\begin{equation}
    \label{eq:position_constraint}
    f_c(\bm{q}) = 0,
\end{equation}
where $f_c: \calQ \mapsto \RR^m$ is the constraint function of dimension $m$.
Such constraints act on the system via the constraint forces $\bm{\lambda} \in \RR^m$ that are spanned by the transpose of its Jacobian, denoted by $J_c \in \RR^{m\times n_v}$.
Thus, the constrained equation of motion reads
\begin{equation}
    \label{eq:constrained_motion}
    M(\bm{q})\bm{\dot{v}} +b(\bm{q},\bm{v}) = \bm{\tau} + J_c^\top \bm{\lambda}.
\end{equation}
In order to simulate the constrained system, given a torque input $\bm{\tau}$ and knowing $\bm{q}$, $\bm{v}$, one needs to solve \eqref{eq:constrained_motion} and \eqref{eq:position_constraint} for the unknowns $\bm{\dot{v}}$ and $\bm{\lambda}$.
To do so, it is more convenient to first rewrite \eqref{eq:position_constraint} as a function of $\bm{\dot{v}}$ by proceeding to index reduction.
Thus, differentiating \eqref{eq:position_constraint}
twice with respect to time yields
\begin{equation}
    \label{eq:acc_constraint}
    J_c \bm{\dot{v}} + \underbrace{ \dot{J_c} \bm{v}}_{\gamma(\bm{q},\bm{v})} = 0.
\end{equation}
For index reduction to be valid, it implicitly assumes that the preceding once differentiated and the original systems \eqref{eq:position_constraint} are verified at the current time-step.
Due to numerical reasons, such a condition is never exactly met, and a corrective term is added to $\gamma$ to avoid numerical drift as done in the Baumgarte stabilization techniques.
Combining the equations \eqref{eq:constrained_motion} and \eqref{eq:acc_constraint} leads to
\begin{equation}
    \label{eq:constrained_system}
    \begin{pmatrix}
         M & J_c^\top \\
        J_c & 0_{m\times m}
    \end{pmatrix}
    \begin{pmatrix}
        \bm{\dot{v}} \\
        \bm{-\lambda}
    \end{pmatrix}
     = \begin{pmatrix}
         \bm{\tau} - b(\bm{q},\bm{v}) \\
         - \gamma(\bm{q},\bm{v})
     \end{pmatrix}.
\end{equation}
At this stage, it is worth noting that \eqref{eq:constrained_system} is often not invertible as cases where $J_c$ is rank deficient i.e. hyper-static systems, are common in robotics.

\subsection{Compliant constraints}

In the previously described constrained dynamics, rigid constraints were modeled via \eqref{eq:position_constraint} which enforces $f_c$ to be null.
When modeling compliant constraints, the function $f_c$ is allowed to be non-null but now defines a potential energy $U(\bmq) =\frac{1}{2} \| f_c(\bmq) \|^2_{R^{-1}}$ where $\|x\|_A = \sqrt{x^\top A x}$ for any $x\in \RR^m$ and any $A \in \RR^m$ positive semi-definite.
Here, $R\in \RR^{m\times m}$ denotes compliance, which is a physical property of the system.
Such potential energy leads to a constraint torque $\bm{\tau}_c$
\begin{equation}
    \bm{\tau}_c = - \nabla_{\bmq} U(\bmq) = - J_c^\top \underbrace{R^{-1} f_c(\bmq)}_{-\bm{\lambda}},
\end{equation}
which induces \eqref{eq:position_constraint} to be replaced by a linear mapping between constraint forces and constraint violation
\begin{equation}
    \label{eq:compliant_position_constraint}
    f_c(\bmq) = -R \bm{\lambda}.
\end{equation}
As previously described, proceeding to index reduction with \eqref{eq:compliant_position_constraint} and adding \eqref{eq:constrained_motion}, simulation of compliant constraints leads to the following system
\begin{equation}
    \label{eq:compliant_constrained_system}
    \begin{pmatrix}
         M & J_c^\top \\
        J_c & -R
    \end{pmatrix}
    \begin{pmatrix}
        \bm{\dot{v}} \\
        \bm{-\lambda}
    \end{pmatrix}
     = \begin{pmatrix}
         \bm{\tau} - b(\bm{q},\bm{v}) \\
         - \gamma
     \end{pmatrix},
\end{equation}
whose only difference with \eqref{eq:constrained_system} lies in the null lower right block being replaced by the compliance matrix $R$.
Numerically, this difference appears to be critical.
Indeed, $R$ being positive definite makes the system \eqref{eq:compliant_constrained_system} always well-defined and invertible~\cite{tournier2015stable}.

\subsection{Constrained dynamics from an optimization perspective}

All the previous equations could be alternatively derived from an optimization standpoint.
Indeed, one could observe that \eqref{eq:constrained_system} coincides with the Karush–Kuhn–Tucker (KKT) conditions of the following Quadratic Programming (QP):
\begin{align}
    \label{eq:qp_least_constraint}
    \min_{\bm{\dot{v}}} \ &\frac{1}{2}\| \bm{\dot{v}} - \bm{\dot{v}}_f \|^2_M \\
    \subjto \ &J_c \bm{\dot{v}} + \gamma = 0 \nonumber
\end{align}
where $\bm{\dot{v}}_f = M^{-1}(\bm{\tau} - b(\bm{q},\bm{v}))$ is the so-called joint free acceleration of the unconstrained system \eqref{eq:motion}.
Problem~\eqref{eq:qp_least_constraint} corresponds to the formulation of the least constraint principle~\cite{bruyninckx2000gauss} and is equivalent to the following saddle-point problem:
\begin{equation}
    \label{eq:rigid_saddle}
    \min_{\bm{\dot{v}}} \max_{\bm{\lambda}} \ \mathcal{L}(\bm{\dot{v}}, \bm{\lambda})
\end{equation}
with $\mathcal{L}(\bm{\dot{v}}, \bm{\lambda}) = \frac{1}{2}\| \bm{\dot{v}} - \bm{\dot{v}}_f \|^2_M - \bm{\lambda}^\top(J_c \bm{\dot{v}} + \gamma)$ being the Lagrangian associated to \eqref{eq:qp_least_constraint}.

Similarly, the compliant system \eqref{eq:compliant_constrained_system} could be retrieved from the regularized problem:
\begin{equation}
    \label{eq:compliant_saddle}
    \min_{\bm{\dot{v}}} \max_{\bm{\lambda}} \ \mathcal{L}(\bm{\dot{v}}, \bm{\lambda}) - \frac{1}{2}\|\bm{\lambda}\|^2_R.
\end{equation}
where the deformation energy acts as a Tikhonov regularization on the dual variables (i.e., the constraint forces) of the rigid constrained problem \eqref{eq:rigid_saddle}.

\subsection{Proximal algorithms}

Proximal algorithms~\cite{parikh2014proximal} are a general class of optimization techniques, ubiquitous in convex optimization, and are essentially rooted around the notion of proximal operators.
The proximal operator~\cite{moreau1962fonctions} of a convex function $f$ is defined by
\begin{equation}
    \label{eq:prox_f}
    \bm{\mathrm{prox}}_{\rho, f} (x) = \argmin_y f(y) + \frac{\rho}{2} \| x-y\|_2^2
\end{equation}
where $\rho \in \RR_{>0}$ is homogeneous to the inverse of a step size.
A specific property of this operator is that its fixed points coincide with the set of minimizers of $f$.
The proximal point algorithm \cite{parikh2014proximal} uses the corresponding fixed point iteration
\begin{equation}
    \label{eq:prox_point_algo}
    x^{k+1} = \bm{\mathrm{prox}}_{{\rho},f} (x^k),
\end{equation}
and is known to converge towards a minimizer of $f$.

Considering the particular case of QP, one seeks to solve
\begin{align}
\label{eq:QP}
    \min_{x} \ &\frac{1}{2}x^\top H x + g^\top x \\
    \subjto \ &Ax = b \nonumber
\end{align}
where $H \in \RR^{q\times q}$ is a symmetric positive semi-definite matrix, $A \in \RR^{p\times q}$, $g \in \RR^q$ and $b \in \RR^p$.
The Lagrangian of this problem writes,
\begin{equation}
    \calL (x, z)  = \frac{1}{2}x^\top H x + g^\top x - z^\top (Ax - b),
\end{equation}
where $z \in \RR^p$ is the dual variable.
The solution to the problem verifies
\begin{equation}
    x^*, z^* = \text{arg} \min_x \max_z\calL (x, z)
\end{equation}
and the KKT conditions
\begin{equation}
    \label{eq:kkt_qp}
    \begin{pmatrix}
         H & A^\top \\
        A & 0_{p\times p}
    \end{pmatrix}
    \begin{pmatrix}
        x^* \\
        -z^*
    \end{pmatrix}
     = \begin{pmatrix}
         -g \\
         b
     \end{pmatrix}.
\end{equation}
When applied to the dual part of the QP \eqref{eq:QP}, the proximal point algorithm can be written as
\begin{equation}
    \label{eq:prox_saddle}
    x^{k+1}, z^{k+1} = \text{arg} \min_x \max_z\calL (x, z) - \frac{\rho}{2} \| z - z^k\|_2^2,
\end{equation}
where the \textit{minus sign} in front of the term $- \frac{\rho}{2} \| z - z^k\|_2^2$ comes from the maximization over the dual variable $z$. 
The solution of the proximal iteration defined in Eq.~\eqref{eq:prox_saddle} is obtained by solving the following system of equations
\begin{equation}
    \label{eq:qp_prox_system}
    \begin{pmatrix}
         H & A^\top \\
        A & -\rho \text{Id}
    \end{pmatrix}
    \begin{pmatrix}
        x^{k+1} \\
        -z^{k+1}
    \end{pmatrix}
     = \begin{pmatrix}
         -g \\
         b-\rho z^k
     \end{pmatrix}.
\end{equation}
This approach is widely used in optimization and at the core of several optimization solvers \cite{stellato2020osqp,bambade2022prox} and algorithms~\cite{carpentier2021proximal} for solving constrained dynamical problems.\\

\noindent\textbf{Link between proximal and compliance terms.}
Going back to the simulation of constrained systems, one should notice the similarity between \eqref{eq:compliant_constrained_system} and \eqref{eq:qp_prox_system}.
Interestingly, drawing parallels between these two systems, the proximal regularization acts as a numerical compliance, thus making the iterates \eqref{eq:qp_prox_system} always well-defined.
However, due to different regularization terms in \eqref{eq:compliant_saddle} and \eqref{eq:prox_saddle}, the effect of the proximal operator differs from a standard mechanical compliance.
This translates into a shift in the right-hand side of \eqref{eq:qp_prox_system} through the term $-\rho z^k$.
As further discussed in~\cite{parikh2014proximal}, this term cancels out over the iterations, 
making the proximal solution converge towards the solution of the original rigid problem \eqref{eq:kkt_qp}.
Therefore, proximal regularization can be seen as a numerical technique akin to vanishing compliance.

\subsection{Modeling frictional unilateral contacts: the Nonlinear Complementary Problem}

Previously described equations govern the motion of systems under equality constraints, \textit{i.e.}, bilateral constraints.
However, punctual contact interactions are subject to three main modeling hypotheses: the unilateral contact constraint, the Coulomb friction law, and the Maximum Dissipation Principle (MDP).
Next, we detail these principles and the problem they induce.
\\

\begin{table*}[t]
  \caption{Characteristics of the contact solvers in robotics.}
  \label{tab:contact_solvers}
  \centering
    \begin{tabular}{r|cccc}
      \toprule
                                       Physics engine  &  Complementarity problem & Contact type   & Algorithm\\
      \midrule
      \rowcolor{pastelblue}
      ODE\cite{ode:2008}, PhysX\cite{physx}, DART \cite{DART}         & LCP  & hard   &  PGS \\
      Bullet\cite{coumans2021}         & NCP  & hard   &  PGS \\
      \rowcolor{pastelblue}
      MuJoCo\cite{todorov2012mujoco}, Drake \cite{drake} & CCP & soft & non-smooth Newton \\
      RaiSim\cite{raisim} & -${}^*$ & hard & per-contact bisection \\
      \rowcolor{pastelblue}
      Dojo\cite{howell2022dojo} & NCP & hard & Interior Point \\
     \textbf{Ours}        & \textbf{NCP}  & \textbf{hard \& soft}  & \textbf{ADMM} \\
      \bottomrule
    \end{tabular}\\
  \vspace{0.5em}
  ${}^*\text{RaiSim does not model the contact problem as a complementarity problem.}$
  \vspace{-0.5em}
\end{table*}

\noindent
\textbf{Unilateral contact hypothesis.}
Through the unilateral hypothesis, one assumes that objects cannot interpenetrate, yielding the following inequality:
\begin{equation}
\label{eq:no_interpenetration}
 \Phi (\bmq)_N \geq 0,
\end{equation}
where $\Phi (\bmq) \in \RR^{3n_c}$ denotes the separation vector \cite{ericson2004real} defined as the minimum norm translation vector putting two shapes at a null distance, $n_c$ being the number of contact points, and the subscripts $N$ and $T$ refer to the normal and tangential indices, respectively.

By duality, such a constraint induces contact forces \mbox{$\bm{\lambda} \in \RR^{3n_c}$} that can only act in a repulsive fashion and when objects are in contact, \textit{i.e.}, when $\Phi (\bmq)_N = 0 $.
These constraints are summarized by the \textit{Signorini condition} \cite{signorini1959questioni}
\begin{equation}
    \label{eq:signorini_condition}
    0 \leq \bm{\lambda}_N \perp \Phi (\bmq)_N \geq 0.
\end{equation}
where $a \perp b$ for vectors $a$ and $b$ means $a^\top b = 0$.

We use an impulse-based formulation to deal with rigid dynamics and impacts.
Applying the Euler symplectic scheme to discretize \eqref{eq:constrained_motion} leads to 
\begin{equation}
    \label{eq:discrete_motion}
    M( \bmv^{t+1} - \bmv_f ) = J_c^\top \bm{\lambda}
\end{equation}
where $\bm{\lambda}$ now denotes an impulse, $\bmv_f = \Delta t \bm{\dot{v}}_f$, $\Delta t$ being the time step, and $J_c \in \RR^{3n_c \times n_v}$ is the Jacobian of $\Phi$.
Here, the choice of integration scheme only requires one costly evaluation of $M$, $J_c$, $\bmv_f$ at the current time step $\bmq^t$, $\bmv^t$.
More involved implicit integrators could be envisaged to improve stability, but they would also need to evaluate these operators multiple times.
In particular, they would require multiple calls to the contact-detection and rigid-body algorithms, which can be computationally expensive.
As was done earlier, the \textit{Signorini condition} can be written in impulse via index reduction \cite{moreau1988mdp}
\begin{equation}
    \label{eq:impulse_signorini_condition}
    0 \leq \bm{\lambda}_N \perp (J_c \bmv^{t+1}+ \gamma)_N \geq 0
\end{equation}
where $\gamma$ is equal to $\Phi(q^t)$ plus previously mentioned corrective terms. 
We note $\bm{c} = J_c \bmv^{t+1} \in \RR^{3n_c}$ the contact point velocities.

Similarly to the bilateral case from Sec.~\ref{sec:background}, any contact constraint can be made compliant by modifying the \textit{Signorini condition}
\begin{equation}
    \label{eq:compliant_signorini_condition}
    0 \leq \bm{\lambda}_N \perp (J_c \bmv^{t+1}+ \gamma + R \bm{\lambda})_N\geq 0.
\end{equation}
The additional compliance term of \eqref{eq:compliant_signorini_condition} introduces possible interpenetration which can be used to model soft bodies. 
More involved deformation models based on the finite element method \cite[]{reddy1993introduction} are also used in robotics \cite[]{ma2019dense} but requires an extra computational cost.\\

\noindent
\textbf{Coulomb's law of friction} is generally employed to model dry friction via
\begin{equation}
    \label{eq:coulomb}
    \forall i, \ \lambda^{(i)} \in \calK_{\mu^{(i)}}
\end{equation}
where \mbox{$\calK_{\mu^{(i)}} =   \{ \lambda^{(i)} \in \RR^3 | \  \|\lambda^{(i)}_T \|_2 \leq \mu^{(i)} \lambda_N^{(i)}\} $} is the second-order friction cone, \mbox{$\mu = \begin{pmatrix}
    \mu^{(1)}, \dots, \mu^{(n_c)} 
\end{pmatrix} \in \RR_{>0}^{n_c}$} is the vector of coefficient of frictions 
and the superscript $(i)$ refers to the indices associated with the i-th contact point.
By noting the Cartesian product \mbox{$\calK_\mu = \prod_{i=1}^{n_c} \calK^{(i)}$}, \eqref{eq:coulomb} can be aggregated into \mbox{$\bm{\lambda} \in \calK_\mu$}. \\

\noindent
\textbf{Maximum dissipation principle.}
In addition, the maximum dissipation principle (MDP) \cite{moreau1988mdp} states that, whenever a contact point is sliding, friction impulses should maximize the dissipated power
\begin{align}
    \label{eq:mdp}
    \forall i, \ \bm{\lambda}_T^{(i)} = &\argmin_{\beta_T, \|\beta_T\| \leq \mu^{(i)} \bm{\lambda}_N^{(i)}} \beta_T^\top \bm{c}_T^{(i)} .
\end{align}
where $c$ is the contact point velocity previously defined.

\noindent
\textbf{Nonlinear complementary problem.}
Finally, combining equations \eqref{eq:discrete_motion}, \eqref{eq:compliant_signorini_condition}, \eqref{eq:coulomb}, \eqref{eq:mdp}, $\bm{\lambda}$ verifies the following NCP
\begin{align}
    \label{eq:NCP}
    \mathcal{K}_\mu &\ni \bm{\lambda} \perp \bm{\sigma} + \Gamma (\bm{\sigma}) \in \mathcal{K}_\mu^*
\end{align}
where \mbox{$\calK_\mu^* = \calK_{\frac{1}{\mu}}$} is the dual friction cone of $\calK_{\mu}$, 
\mbox{$G = J_c M^{-1} J_c^\top$} is the so-called Delassus matrix~\cite{delassus1917memoire}, 
\mbox{$g = J_c \bmv_f + \gamma$} is the free contact point velocities plus corrective terms,
and we use the shorthand notation $\bm{\sigma} = (G+R)\bm{\lambda} + g $.
We use the notation 
\mbox{$\Gamma(\bm{\sigma}) =\begin{pmatrix}
    \Gamma^{(1)}(\bm{\sigma}^{(1)}) &  \dots &   \Gamma^{(n_c)}(\bm{\sigma}^{(n_c)})
\end{pmatrix}\in \RR^{3n_c}$} with \mbox{$\Gamma^{(i)}(\bm{\sigma}^{(i)}) =\begin{pmatrix}
    0 &  0 &   \mu^{(i)} \| \bm \sigma_T^{(i)} \|
\end{pmatrix}\in \RR^{3}$} denoting the De Saxc\'e correction~\cite{desaxce1998bipotential}.

\subsection{Existing solvers}

Due to the non-convexity and nonsmoothness of the complementarity constraint, the nonlinearity of the DeSaxc\'e correction, and the ill-conditioning of the Delassus matrix $G$, the NCP~\eqref{eq:NCP} is known to be a numerically hard problem to solve in general~\cite{acary2017contact}.

A first class of simulators, e.g., ODE~\cite{ode:2008}, PhysX~\cite{physx}, DART\cite{DART}, proceed by linearizing the friction cones to get an approximate but more tractable Linear Complementarity Problem (LCP).
LCPs are well-studied~\cite{cottle2009lcp} and can be solved with the Projected Gauss-Seidel (PGS) algorithm.
As a first-order algorithm, PGS is not affected by null eigenvalues due to hyperstaticity but is sensitive to the conditioning of the Delassus matrix $G$, thus hindering robustness.
In its recent versions, Bullet~\cite{coumans2021} implements a similar PGS algorithm working on the original second-order friction cone.

An alternative approach, notably adopted in MuJoCo\cite{todorov2012mujoco} and Drake\cite{drake}, consists in ignoring the DeSaxc\'e correction in \eqref{eq:NCP}, which has the effect of relaxing the \textit{Signorini condition}~\cite{lelidec2023contact}.
This leads to a Cone Complementarity Problem (CCP) which is equivalent to a convex Second-Order Cone Programming (SOCP) problem~\cite{boyd2004convex}.
Such a problem can be solved via robust off-the-shelf optimization algorithms benefiting from strong guarantees.
In this respect, a previous work \cite{tasora2021solving} uses an ADMM algorithm to solve this problem.
In MuJoCo~\cite{todorov2012mujoco} and Drake~\cite{drake} simulators, the hyperstatic cases are handled by systematically adding compliance to the problem, thus making it impossible to simulate purely rigid systems.

Raisim~\cite{raisim} uses a specific contact model enforcing the \textit{Signorini condition}, which combines favorably with its custom per-contact bisection algorithm. 
Despite its computational efficiency, this approach inherits the sensitivity to conditioning from Gauss-Seidel-like techniques.

Dojo \cite{howell2022dojo} avoids any physical relaxation and uses an Interior Point (IP) algorithm to solve \eqref{eq:NCP}.
As a second-order algorithm, Dojo's solver can handle ill-conditioned problems.
However, just like IP algorithms, the resulting algorithm is difficult to warm-start and requires an expensive Choleksy computation at each iteration, which makes the approach time-consuming, and thus limits its range of applications, notably in the context of real-time control scenarios.

\section{Efficient solving of frictional contact dynamics} 
\label{sec:admm}

In this section, we detail the central contribution of this paper, namely a novel algorithm to solve NCPs of the form of problem~\eqref{eq:NCP}. 
At the core of our approach, is the development of a new ADMM method and update strategy, enabling us to solve complex contact problems, that might be poorly conditioned, as it might occur in contact mechanics problems. 

\subsection{NCP as a cascade of optimization problems}

The NCP formulated in~\eqref{eq:NCP} corresponds to the solving of the interweaving problems of the form
\begin{subequations}
\label{eq:qcqp_FD}
\begin{align}
    \bm{\lambda} &=
    \argmin_{\bm{f}\in \calK_\mu}
    \frac{1}{2} \bm{f}^\top (G + R) \bm{f} 
    + \bm{f}^\top \left(\Gamma(\bm\sigma) + g\right)\,, \label{subeq:qcqp_FD_primal_update} \\
    \text{and}& \nonumber \\
    \bm{\sigma} &= (G+R) \bm{\lambda} + g\,. \label{subeq:qcqp_FD_dual_update}
\end{align}
\end{subequations}
Indeed, if we denote by $\bm{z} \in \calK^{*}_\mu$ the dual variable associated to the primal variable $\bm{\lambda} \in \calK_\mu$, the Lagrangian of Problem~\eqref{eq:qcqp_FD} reads
\begin{equation}
    \mathcal{L}(\bm{f},\bm{z}) = \frac{1}{2} \bm{f}^\top (G + R) \bm{f} 
    + \bm{f}^\top \left(\Gamma(\bm\sigma) + g\right)
    - \bm{f}^\top \bm{z}.
\end{equation}
The optimality conditions are thus given by canceling the gradient of $\mathcal{L}$ w.r.t. $\bm f$ at the optimal solution $(\bm \lambda, \bm{z}^*)$
\begin{equation}
    \label{eq:nabla_L}
    \nabla_{\bm f} \mathcal{L} (\bm \lambda, \bm{z}^*) = (G + R) \bm{\lambda} + g + \Gamma(\bm\sigma) - \bm{z}^* = 0,
\end{equation}
and the fact that
\begin{equation}
    \label{eq:optimality_subproblem}
    \calK_\mu \ni \bm \lambda \perp \bm{z}^*  \in \calK_\mu^{*}.
\end{equation}
Injecting \eqref{eq:nabla_L} into \eqref{eq:optimality_subproblem}, leads to the nonlinear complementarity conditions of~\eqref{eq:NCP}.

Interestingly, \eqref{subeq:qcqp_FD_primal_update} appears to be a convex problem which proves to be useful as it allows sub-steps of our ADMM approach to be performed efficiently.
However, the interweaving of \eqref{subeq:qcqp_FD_primal_update} and \eqref{subeq:qcqp_FD_dual_update} through the non-smooth de Saxc\'e function $\Gamma$ makes the problem~\eqref{eq:NCP} difficult to solve in general by standard optimization techniques~\cite{acary2017contact}.
As discussed in~\cite{acary2017contact}, one approach consists in incorporating estimates of $\Gamma(\bm \sigma)$ (which is equal to $\Gamma(\bm z^*)$), via the updates of a variable named $s$, inside the solving of \eqref{subeq:qcqp_FD_primal_update}.
More specifically, this results in a cascade of optimization problems and is done by setting, at the $k^\text{th}$ iterate, $s_k = \Gamma(\bm z_k)$ (Alg.~\ref{alg:admm_ncp}, line~\ref{line:s_update}), corresponding to the nonlinear term in~\eqref{eq:qcqp_FD} considered as constant, then solving~\eqref{subeq:qcqp_FD_primal_update} (Alg.~\ref{alg:admm_ncp}, line~\ref{line:primal_update1}) and updating $\bm z_k$ (Alg.~\ref{alg:admm_ncp}, line~\ref{line:dual_update}) from the new optimal force vector $\bm f_{k}$.
The solving of this cascade of problems continues until the primal and dual optimal convergence criteria have been met up to a certain numerical tolerance $\epsilon_\text{abs}$.

This cascaded strategy has shown to be effective in practice~\cite{acary2017contact}. 
A critical part involves solving the inner optimization problem~\eqref{subeq:qcqp_FD_primal_update} efficiently, which is nothing more than a Quadratically Constrained Quadratic Program (QCQP).
One approach could thus consist in leveraging existing SOCP solvers, such as ECOS~\cite{domahidi2013ecos} or SCS~\cite{odonoghue:21}.
Yet, because of their high level of versatility, such off-the-shelf solvers tend to be less efficient when considering a specific class of problems such as the ones occurring in contact mechanics~\cite{acary2017contact}.
Following this line of thought, we instead propose to develop a dedicated and efficient ADMM approach for solving~\ref{subeq:qcqp_FD_primal_update}, particularly suited for solving ill-conditioned problems.

\subsection{Proximal ADMM formulation}

By introducing a slack variable $\bm y$ such that $\bm y = \bm f$, the QCQP \eqref{subeq:qcqp_FD_primal_update} is made separable as follows
\begin{subequations}
    \label{eq:split_QCQP}
    \begin{align}
    \min_{\bm f, \bm y} \ & h_1(\bm f) + h_2(\bm y) \\
    \subjto \ & \bm f = \bm y \label{subeq:split_QCQP_equality} \,, 
    \end{align}
\end{subequations}
where $h_1(\bm f) := \frac{1}{2} \bm{f}^\top (G+R) \bm{f} + \bm{f}^{\top}(s + g) $ is a smooth convex function with $s$ being the current estimate of $\Gamma(\bm \sigma)$, and $h_2(\bm y) := \mathcal{I}_{\calK_\mu}(\bm y)$ is the nonsmooth indicator function associated with the convex cone $\calK_\mu$ defined as
\[
    \mathcal{I}_{\calK_\mu}(\bm y)= 
\begin{cases}
    0,& \text{if } \bm y \in \calK_\mu\\
    +\infty,              & \text{otherwise}
\end{cases}.
\]

By naming $\bm z$ the dual variable associated with \eqref{subeq:split_QCQP_equality}, the augmented Lagrangian of problem~\eqref{eq:split_QCQP} reads
\begin{equation}
\label{eq:AL_ADMM}
    \mathcal{L}_\rho^A(\bm f, \bm y, \bm z) = 
    h_1(\bm f) + h_2(\bm y) 
    - \bm{z}^{\top}(\bm f - \bm y) 
    + \frac{\rho}{2} \| \bm f - \bm y \|^2_2\,
\end{equation}
where $\rho$ is the augmented Lagrangian penalty term~\cite{bertsekas2014constrained}.
$\mathcal{L}_\rho^A$ defined in~\eqref{eq:AL_ADMM} can be equivalently rewritten as
\begin{equation}
\label{eq:AL_ADMM_2}
    \mathcal{L}_\rho^A(\bm f, \bm y, \bm z) = 
    h_1(\bm f) + h_2(\bm y)
    + \frac{\rho}{2} \left\| \bm f - \bm y - \frac{\bm{z}}{\rho} \right\|^2_2\
    - \frac{1}{\rho} \left\| \bm z \right\|^2_2 \,.
\end{equation}

Finally, to make the sub-problem associated to $h_1$ strongly convex w.r.t. $\bm f$, we propose to add the proximal term \mbox{$\frac{\eta}{2} \| \bm f - \bm{f}^{-} \|^2_2$} to $\mathcal{L}_\rho^A$, leading to
\begin{multline}
\label{eq:AL_ADMM_prox}
    \mathcal{L}_{\rho,\eta}^A(\bm f, \bm y, \bm z) 
    = h_1(\bm f) + \frac{\eta}{2} \| \bm f - \bm{f}^{-} \|^2_2
    + h_2(\bm y) \\
    + \frac{\rho}{2} \left\| \bm f - \bm y - \frac{\bm{z}}{\rho} \right\|^2_2
    - \frac{1}{\rho} \left\| \bm z \right\|^2_2,
\end{multline}
where $\bm{f}^{-}$ is the previous estimate of $\bm{f}$.
This proximal term affects the conditioning of the problem by adding a regularization to the Hessian of $h_1$, which will play an essential role in the update strategy further detailed in Sec.~\ref{subsec:admm_update_parameter}.
$\eta$ is fixed and typically set to a value of $10^{-6}$.

\subsection{Pseudocode of the ADMM updates}

\begin{figure}
    \centering
    \hspace{-0.2cm}
    \includegraphics[width=0.48\linewidth]{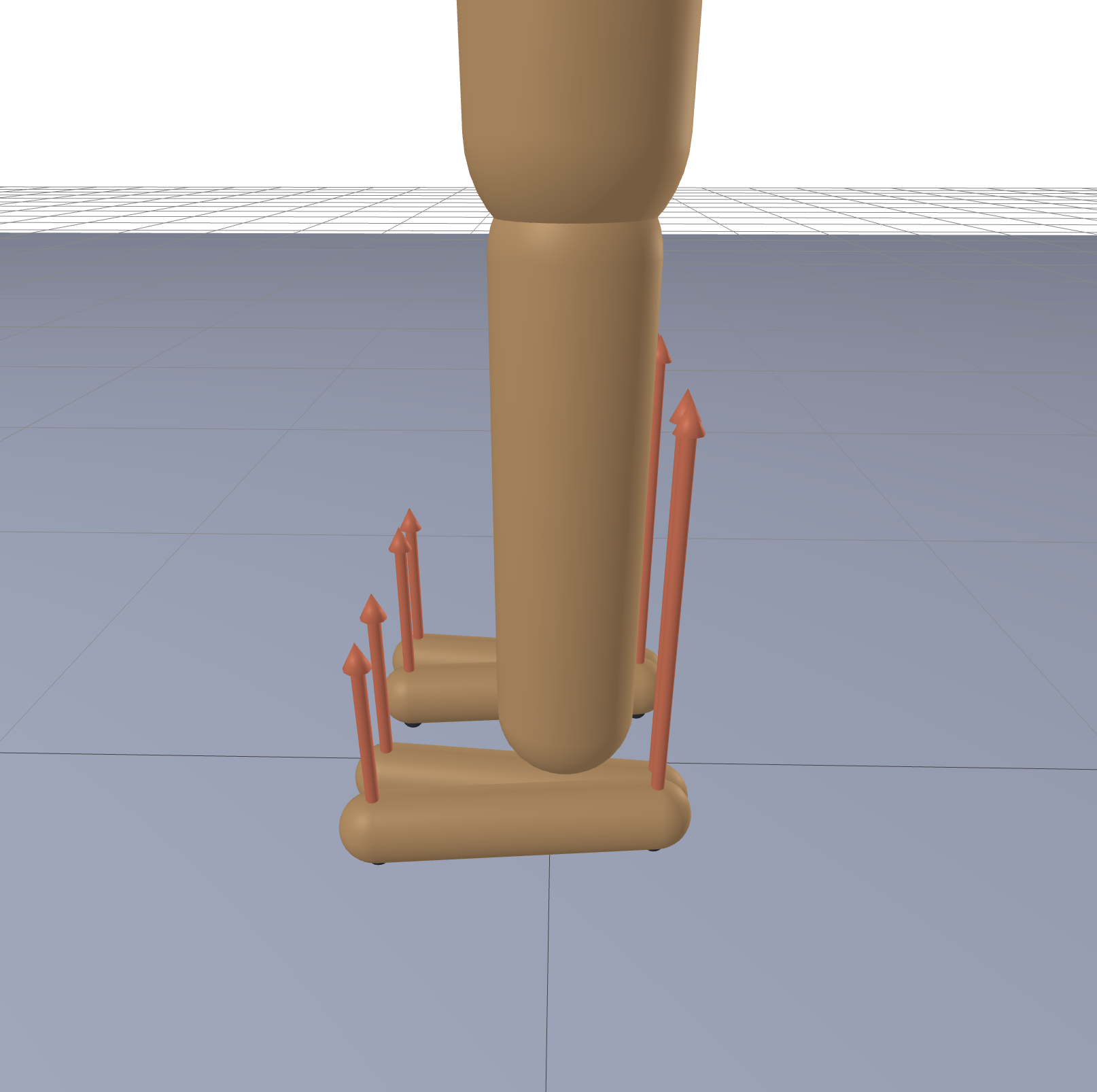}
    \includegraphics[width=0.48\linewidth]{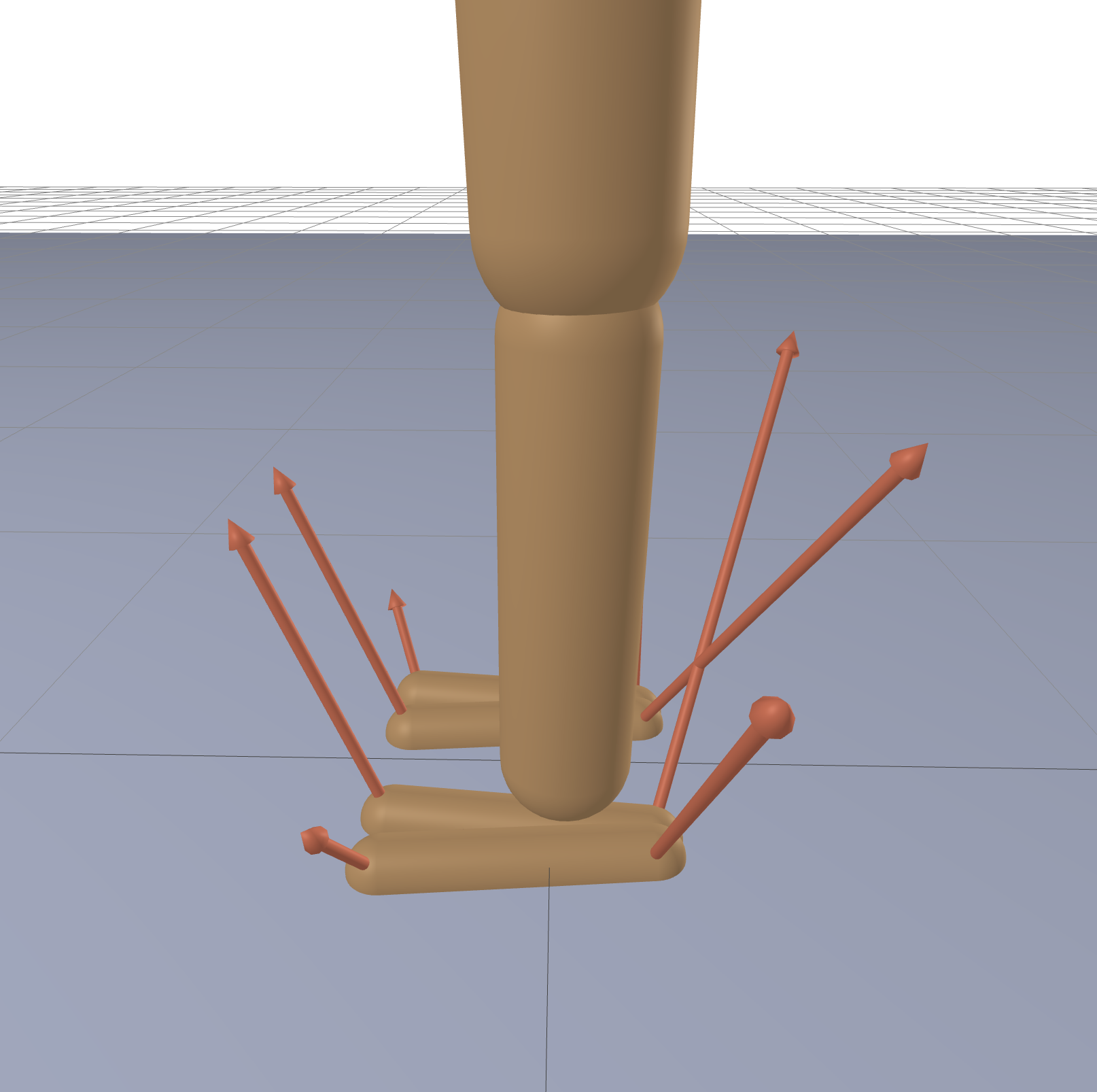}
    \caption{
    \textbf{Compliant contacts}.
    Our contact solver can handle both purely rigid (\textbf{left}) and compliant contacts (\textbf{right}).
    }
    \label{fig:humanoid_compliant_contact}
\end{figure}

Solving the saddle-point point problem associated with the proximal augmented Lagrangian $\mathcal{L}_{\rho,\eta}^A$ defined in~\eqref{eq:AL_ADMM_prox} can be efficiently done via ADMM iterates.
The updates are defined by the following successive optimization steps:
\begin{subequations}
    \label{eq:ADMM_pseudocode}
    \begin{align}
        \bm{f}_{k} &= \argmin_{\bm{f}} \mathcal{L}_{\rho,\eta}^A(\bm f, \bm y_{k-1}, \bm z_{k-1}) \label{subeq:ADMM_pseudocode1} \\ 
        \bm{y}_{k} &= \argmin_{\bm{y}} \mathcal{L}_{\rho,\eta}^A(\bm f_{k}, \bm y, \bm z_{k-1}) \label{subeq:ADMM_pseudocode2} \\ 
        \bm{z}_{k} &= \bm{z}_{k-1} - \rho (\bm f_{k} - \bm{y}_{k}) \label{subeq:ADMM_pseudocode3} \,. 
    \end{align}
\end{subequations}
We now detail the content of each sub-computations of~\eqref{eq:ADMM_pseudocode}.
The first equation~\eqref{subeq:ADMM_pseudocode1} corresponds to the solving of an unconstrained quadratic problem, whose solution is given by:
\begin{multline}
    \label{subeq:ADMM_pseudocode1_detailed}
    \bm{f}_{k} = -\left(G+R+(\eta + \rho) \text{Id}\right)^{-1} \\
    \left( g + s_k -\eta \bm{f}_{k-1} - \rho \bm{y}_{k-1} - \bm{z}_{k-1} \right) \,.
\end{multline}
It is worth noting that \eqref{subeq:ADMM_pseudocode1_detailed} is always well-defined because of the proximal regularization.
This allows our ADMM approach to handle both compliant and purely rigid contacts as illustrated in Fig.\ref{fig:humanoid_compliant_contact}.
The second equation~\eqref{subeq:ADMM_pseudocode2} can be explicitly written as:
\begin{equation}
    \bm{y}_{k} = \argmin_{\bm{y} \in \calK_\mu} \frac{\rho}{2} \left\| \bm f_{k}  - \frac{\bm{z}_{k-1}}{\rho} - \bm y \right\|^2_2 \,.
\end{equation}
It corresponds to the orthogonal projection of the vector $\bm f_{k}  - \frac{\bm{z}_{k-1}}{\rho}$ on the Cartesian product of friction cones $\calK_\mu$, also denoted by:
\begin{equation}
    \bm{y}_{k} = \calP^\text{Id}_{\calK_\mu}\left( \bm f_{k}  - \frac{\bm{z}_{k-1}}{\rho} \right)
\end{equation}
The third equation~\eqref{subeq:ADMM_pseudocode3} corresponds to the classic augmented Lagrangian multiplier updates~\cite{bertsekas2014constrained}.

%

\subsection{Primal and dual convergence criteria}

As classically done in the ADMM settings, the primal and dual residuals associated with the augmented Lagrangian function $\mathcal{L}_{\rho,\eta}^A$ at the $k^\text{th}$ iterates are respectively given by
\begin{equation}
    \label{eq:primal_residual}
    \text{r}_k^\text{prim} = \bm f_k - \bm y_k \,,
\end{equation}
and
\begin{equation}
    \label{eq:dual_residual}
    \text{r}_k^\text{dual} = \eta (\bm f_k - \bm f_{k-1})  + \rho(\bm y_k - \bm y_{k-1})  \,.
\end{equation}
In addition, we use a contact complementarity residual defined as 
\begin{equation}
    \label{eq:comp_residual}
    \text{r}_k^\text{comp} = \begin{pmatrix}
        | \bm {f_k^{(1)\top}} \bm z_k^{(1)}| & \dots & | \bm {f_k^{(n_c)\top}} \bm z_k^{(n_c)}|
    \end{pmatrix}.
\end{equation}
The iterations of ADMM have converged to precision $\epsilon_\text{abs}$ when:
\begin{equation}
    \label{eq:stopping_criteria}
 \|  \text{r}_k^\text{prim} \|_\infty, \|  \text{r}_k^\text{dual} \|_\infty, \|  \text{r}_k^\text{comp} \|_\infty \leq \epsilon_\text{abs}
\end{equation}
A typical set of values for $\epsilon_\text{abs}$ which offers a good compromise between realistic simulation (compared to alternative resolution methods) and computation times to solve the NCP problems is $[ 10^{-4}; 10^{-6} ]$.
We choose to use the infinity norm ($\| . \|_\infty$) as it is independent of the problem dimensions.
As explained in~\cite{parikh2014proximal}, it is also possible to introduce relative convergence criterion, as done in many optimization solvers of the literature~\cite{odonoghue:21,bambade2022prox}, to account for the potential stagnation of the optimization variables due to the numerics. 

\subsection{Exploiting problem sparsity}
\label{subsec:sparsity}
Over the three main steps of the ADMM recursion \eqref{eq:ADMM_pseudocode}, the last two \eqref{subeq:ADMM_pseudocode2} and \eqref{subeq:ADMM_pseudocode3} are cheap operations of linear complexity w.r.t. the problem dimensions.
The most complex operation lies in the resolution of the linear system in Eq.~\eqref{subeq:ADMM_pseudocode1}, and detailed in Eq.~\ref{subeq:ADMM_pseudocode1_detailed}.
It notably requires the inversion of the augmented Delassus matrix, denoted by \mbox{$G_{\rho,\eta}^R := G + R + (\eta + \rho)\text{Id}$} in the sequel.
Interestingly, the additional terms $R + (\eta + \rho)\text{Id}$ only modify the diagonal of the Delassus matrix $G$ with positive elements.
In other words, both the sparsity of the Delassus matrix and the positivity are preserved, allowing to directly call sparse or dense Cholesky methods to decompose the augmented Delassus $G_{\rho,\eta}^R$.

Additionally, in the case of kinematics algorithms composed of multiple joints including loop-closure, one can leverage branch-inducing sparsity algorithms~\cite{featherstone2014rigid} to efficiently evaluate the Delassus matrix~\cite{featherstone2010exploiting,sathya2023pvosimr} or directly obtain its Cholesky decomposition at a reduced cost~\cite{carpentier2021proximal}.
Finally, it is worth noticing that, as soon as the ADMM penalty term $\rho$ is updated, it requires the full Cholesky refactorization of $G_{\rho,\eta}^R$.
In other words, to lower the computational footprint induced by the successive Cholesky factorizations, it is essential to lower the number of updates of $\rho$. 
This motivates the introduction of the spectral update rule for $\rho$, detailed in the next subsection, which is experimentally validated in Sec.~\ref{sec:results}.

\subsection{ADMM parameters update strategies}
\label{subsec:admm_update_parameter}

The convergence rate of the ADMM methods is directly related to the value of the augmented penalty term $\rho$~\cite{parikh2014proximal}.
Choosing this value directly depends on input problem values ($G, R, g$ and $\calK_\mu$).
There is no automatic procedure to choose the best $\rho$ which will lower the number of iterations to reach the desired primal/dual accuracy $\epsilon_\text{abs}$. \\

\noindent
\textbf{Linear update rule.}
From a given initial value $\rho$, a well-known strategy is to linearly update $\rho$ according to the ratio between primal and dual residuals ${\text{r}_k^\text{prim}}/{\text{r}_k^\text{dual}}$, following the update rule at the $k^\text{th}$ iterate
\begin{equation}
        \rho^\text{new} = 
\begin{cases}
    \tau^\text{inc} \rho & \text{if } \|  \text{r}_k^\text{prim} \|_\infty \geq \alpha \|  \text{r}_k^\text{dual} \|_\infty \\
    {\rho}/{\tau^\text{dec}} & \text{if } \|  \text{r}_k^\text{dual} \|_\infty \geq \alpha \|  \text{r}_k^\text{prim} \|_\infty \\
    \rho,              & \text{otherwise}\,,
\end{cases}
\end{equation}
where $\tau^\text{inc} > 1 $ and $\tau^\text{dec} > 1$ are increment/decrement factors, and $\alpha > 1$ is the ratio parameter between primal and dual residuals.
The overall idea is to maintain the trajectory of primal and dual residual norms within a tube of diameter $\alpha$.
Yet, if the problem is poorly conditioned, this linear update rule will often trigger many updates of $\rho$, thus requiring each time to recompute the Cholesky factorization associated with the nonsingular augmented Delassus matrix $G_{\rho,\eta}^R$.\\

\noindent
\textbf{Spectral update rule.}
To overcome the inherent limitations of the standard linear ADMM update rule, we introduce a new update strategy that accounts for the spectral properties of the augmented Delassus matrix $G_{\rho,\eta}^R$. 
More precisely, our approach is inspired by the work of \citet{nishihara2015general} which provides a convergence analysis of a generic class of ADMM formulations, including ours depicted by~\eqref{eq:split_QCQP}.
Their analysis assumes that the ADMM penalty parameter $\rho$ is of the form
\begin{equation}
    \rho := \sqrt{mL} \,\, \kappa^p\,,
\end{equation}
where $m$ is the strong convexity parameter of $h_1$ and $L$ is the Lipschitz constant associated with $\nabla h_1$, where $h_1$ refers to the smooth convex function in the ADMM formulation~\eqref{eq:split_QCQP}. As $h_1$ is a quadratic function, $m$ and $L$ respectively correspond to the lowest and largest eigenvalues of $G_{\rho,\eta}^R$.
The ratio $\kappa := L / m$ is the condition number of $h_1$ and $G_{\rho,\eta}^R$.
$p$ is the free exponent parameter directly balancing the contribution of the condition number $\kappa$ in the choice of $\rho$.

While in~\cite{nishihara2015general} $p$ is assumed to be constant, we suggest adjusting its value according to the ratio between primal and dual residuals ${\|\text{r}_k^\text{prim}}\|_\infty/{\|\text{r}_k^\text{dual}\|_\infty}$, in the spirit of the linear update rule recalled previously.
More precisely, we propose this selection strategy
\begin{align}
    \label{eq:spectral_rule}
        p^\text{new} &= 
\begin{cases}
    p + p^\text{inc} & \text{if } \|  \text{r}_k^\text{prim} \|_\infty \geq \alpha \|  \text{r}_k^\text{dual} \|_\infty, \\
    p - p^\text{dec} & \text{if } \|  \text{r}_k^\text{dual} \|_\infty \geq \alpha \|  \text{r}_k^\text{prim} \|_\infty, \\
    p,              & \text{otherwise}\,,
\end{cases}\\
\rho^\text{new} &= \sqrt{mL} \,\, \kappa^{p_\text{new}}, \nonumber
\end{align}
where $p^\text{inc}$ and $p^\text{dec}$ are increments on the exponent parameter $p$.
A typical value is \mbox{$p^\text{inc} = p^\text{dec} = 0.05$}.
As for the linear update rule, $\alpha > 1$ is the ratio parameter between primal and dual residuals, forcing the primal and dual residual norms to lie within a tube of diameter $\alpha$.
To the best of the author's knowledge, this spectral update strategy is novel and directly scales the ADMM update parameter according to the smoothness of the NCP problem.

It is worth noticing that the proposed solution only considers the lowest and largest eigenvalues of $G_{\rho,\eta}^R$, which can be easily estimated from the power iteration algorithm for instance, which, in practice, converges in very few iterations compared to the problem dimensions.
Finally, thanks to the presence of the proximal term added in \eqref{eq:AL_ADMM_prox}, we have $m \geq \eta > 0$, which guarantees the well-posedness of the strategy.

\subsection{Pseudocode}

Algorithm~\ref{alg:admm_ncp} summarizes our ADMM-based approach for solving the NCP~\eqref{eq:NCP} problem of frictional contacts simulation.
It takes as inputs the contact problem parameters, such as the Delassus matrix $G$,  the free contact point velocities $g$, the Cartesian product of the friction cones $\calK_\mu$, the compliance matrix $R$, as well as a desired precision $\epsilon_\text{abs}$.
The outputs of Alg.~\ref{alg:admm_ncp} at line~\ref{line:algo_output} correspond to both the optimal contact forces $\bm{\lambda}$ and the contact point velocities $\bm{\sigma}$.
This last quantity can be directly obtained from the dual variable $\bm{z}$ of the NCP problem, which corresponds to the sum of the contact point velocity $\bm{\sigma} = (G+R) \bm{\lambda} + g$ and the DeSax\'e corrective term $\Gamma(\bm \sigma)$.

\begin{algorithm}
    \SetAlgoLined
    \KwIn{Delassus matrix $G$,  free contact point velocities $g$, friction cones $\calK_\mu$, compliance $R$, desired precision $\epsilon_\text{abs}$.}
    \KwOut{Contact impulses $ \bm{\lambda}$ and contact velocities $ \bm{\sigma}$}
    
    \For{$k=1$ \KwTo $n_{\text{iter}}$}{
        \tcc{ADMM updates}
        $s_k \leftarrow \Gamma(\bm{z}_{k-1})$ \; \label{line:s_update}
        \nosemic $\bm{f}_{k} \leftarrow -(G+R+(\eta + \rho) I_d)^{-1}$\;
          \pushline\dosemic $\left(g + s_k -\eta \bm{f}_{k-1} - \rho \bm{y}_{k-1} - \bm{z}_{k-1} \right) $ \; \label{line:primal_update1}
        \popline $ \bm{y}_{k} \leftarrow \calP^\text{Id}_{\calK_\mu}\left( \bm f_{k}  - \frac{\bm{z}_{k-1}}{\rho} \right)$\; \label{line:primal_update2}
        $\bm{z}_{k} \leftarrow \bm{z}_{k-1} - \rho \left(\bm{f}_{k} - \bm{y}_{k}\right)$\; \label{line:dual_update}
        \tcc{Primal/dual criteria evaluation}
        compute the primal/dual convergence criteria given by \eqref{eq:primal_residual}, \eqref{eq:dual_residual} and the stopping criteria \eqref{eq:stopping_criteria}\;
        \If{converged}{
            break\;
        }
        \tcc{Update $\rho$}
        Update $\rho$ according to the spectral strategy described in Sec.~\ref{subsec:admm_update_parameter}.\;
    }
    $ \bm{\lambda} \leftarrow \bm{y}_k $ and $\bm{\sigma} \leftarrow \bm{z}_k - \Gamma(\bm{z}_k ) $\; \label{line:algo_output}
    \caption{Pseudocode of the ADMM algorithm for NCP for rigid and compliant contacts.}
    \label{alg:admm_ncp}
\end{algorithm}

\section{Inverse Dynamics} 
\label{sec:inverse}

We now consider the inverse dynamics problem, corresponding to the search of the torque $\bm{\tau}$ and the contact impulses $\bm{\lambda}$ that induce a given joint velocity $ \bmv_{\text{ref}}$.
Starting from the NCP~\eqref{eq:NCP} formulation and enforcing the contact point velocities to be equal to $J_c \bmv_{\text{ref}}$ yields the inverse dynamics problem,
\begin{align}
    \label{eq:NCP_ID}
    \mathcal{K}_\mu &\ni \bm{\lambda} \perp \bm{\sigma} + \Gamma (\bm{\sigma}) \in \mathcal{K}_\mu^* \\
    \bm{\sigma} &= R\bm{\lambda} + J_c \bmv_{\text{ref}} + \gamma \nonumber.
\end{align}
We recall that, from KKT conditions, a solution of \eqref{eq:NCP_ID} should minimize
\begin{align}
    \label{eq:qcqp_ID}
    \min_{\lambda \in \calK_\mu}\frac{1}{2} \|\bm{\lambda} + R^{-1}(J_c\bmv_{\text{ref}} +\gamma + s) \|_R^2
\end{align}
where $ s = \Gamma(\bm{\sigma})$.
This corresponds to the projection on $\calK_\mu$ under the metric induced by $R$, so we note \mbox{$\mathcal{P}_{\calK_\mu}^{R}(- R^{-1}(J_c\bmv_{\text{ref}} +\gamma + s))$} the minimizer of \eqref{eq:qcqp_ID}.
We observe the formulation of \eqref{eq:qcqp_ID} becomes ill-defined for the purely rigid case $R = 0$. 
Indeed, as described previously (Sec.~\ref{sec:background}), the rigidity often makes the problem of contact impulses under-determined and thus non-invertible.
As was done in the case of forward dynamics (Sec.~\ref{sec:admm}), we aim to preserve the rigid contact hypothesis by leveraging proximal optimization.
For the case of inverse dynamics, iterating the proximal operator associated to \eqref{eq:qcqp_ID} (Alg.~\ref{alg:prox_ID}, line \ref{line:prox_proj}) allows to find a $\bm{\lambda}$ verifying \eqref{eq:NCP_ID} even in the rigid case.

At this stage, it is worth noting that \eqref{eq:NCP_ID} has the same structure as \eqref{eq:NCP} and is also an NCP.
Therefore, one could have used the algorithm introduced in Sec.~\ref{sec:admm} (Alg.~\ref{alg:admm_ncp}) to solve it.
However, one notable difference between \eqref{eq:NCP} and \eqref{eq:NCP_ID} is the absence of the Delassus matrix in the latter, which has been replaced by the compliance matrix $R$.
Most often, the compliance matrix has a diagonal structure of the form \mbox{$R = \text{Diag}(R^{(1)}_T, R^{(1)}_T, R^{(1)}_N, \dots, R^{(n_c)}_T, R^{(n_c)}_T, R^{(n_c)}_N)$} which can be exploited to design a more efficient algorithm.
Indeed, in the case of a diagonal matrix \mbox{$D \in \RR^{3n_c\times 3n_c}$}, the operator \mbox{$\mathcal{P}_{\calK_\mu}^{D}$} can be computed analytically via the equality \mbox{$\mathcal{P}_{\calK_\mu}^{D} (x)  = D^{-\frac{1}{2}}\mathcal{P}_{\calK_{\tilde{\mu}}}^{\text{Id}}(D^{\frac{1}{2}}x)$} with \mbox{$\tilde{\mu} = \begin{pmatrix}
    \sqrt{\frac{D^{(1)}_T}{D^{(1)}_N}}\mu^{(1)} & \dots & \sqrt{\frac{D^{(n_c)}_T}{D^{(n_c)}_N}}\mu^{(n_c)} 
\end{pmatrix}$} which makes the iterations of Alg.~\ref{alg:prox_ID} computationally cheap.

Once contact forces are determined, the torque $\bm{\tau}$ can be retrieved with a call to the Recursive Newton-Euler Algorithm (RNEA) (Alg.~\ref{alg:prox_ID}, line \ref{line:rnea}).
It is worth noticing that the contribution of the contact torque \mbox{$\bmtau_c =J_c^\top \bm{\lambda}$} can be accounted for through the backward sweep of the RNEA algorithm, thus efficiently exploiting the sparsity induced by the kinematic tree, in the spirit of rigid-body dynamics algorithms~\cite{featherstone2014rigid}.
\begin{algorithm}
    \SetAlgoLined
    \KwIn{Joint velocity $\bmv_{\text{ref}}$, compliance $R$, contact Jacobian $J_c$, friction cones $\calK_\mu$}
    \KwOut{Torque $\bm{\tau}$, contact impulses $ \bm{\lambda}$}
    \For{$k=1$ \KwTo $n_{\text{iter}}$}{
        $s \leftarrow \Gamma(R \bm{\lambda} + J_c \bmv_{\text{ref}} +\gamma)$ \; \label{line:desaxce_id}
       $\bm{\lambda} \leftarrow \mathcal{P}_{\mathcal{K}_\mu}^{R+\rho \text{Id}}(-\big ( R + \rho \text{Id} \big )^{-1}(J_c\bmv_{\text{ref}} +\gamma + s - \rho \bm{\lambda}))$\; \label{line:prox_proj}
    }
    $\bm{\tau} \leftarrow \text{RNEA}(\bmq, \bmv, \bmv_{\text{ref}})- J_c^\top \bm{\lambda}$ \; \label{line:rnea}
    \caption{Pseudo-code of the proximal algorithm for Inverse Dynamics.}
    \label{alg:prox_ID}
\end{algorithm}

\section{Evaluations and Benchmarks}
\label{sec:results}

\begin{figure*}
    \centering
    \includegraphics[width=.3\linewidth]{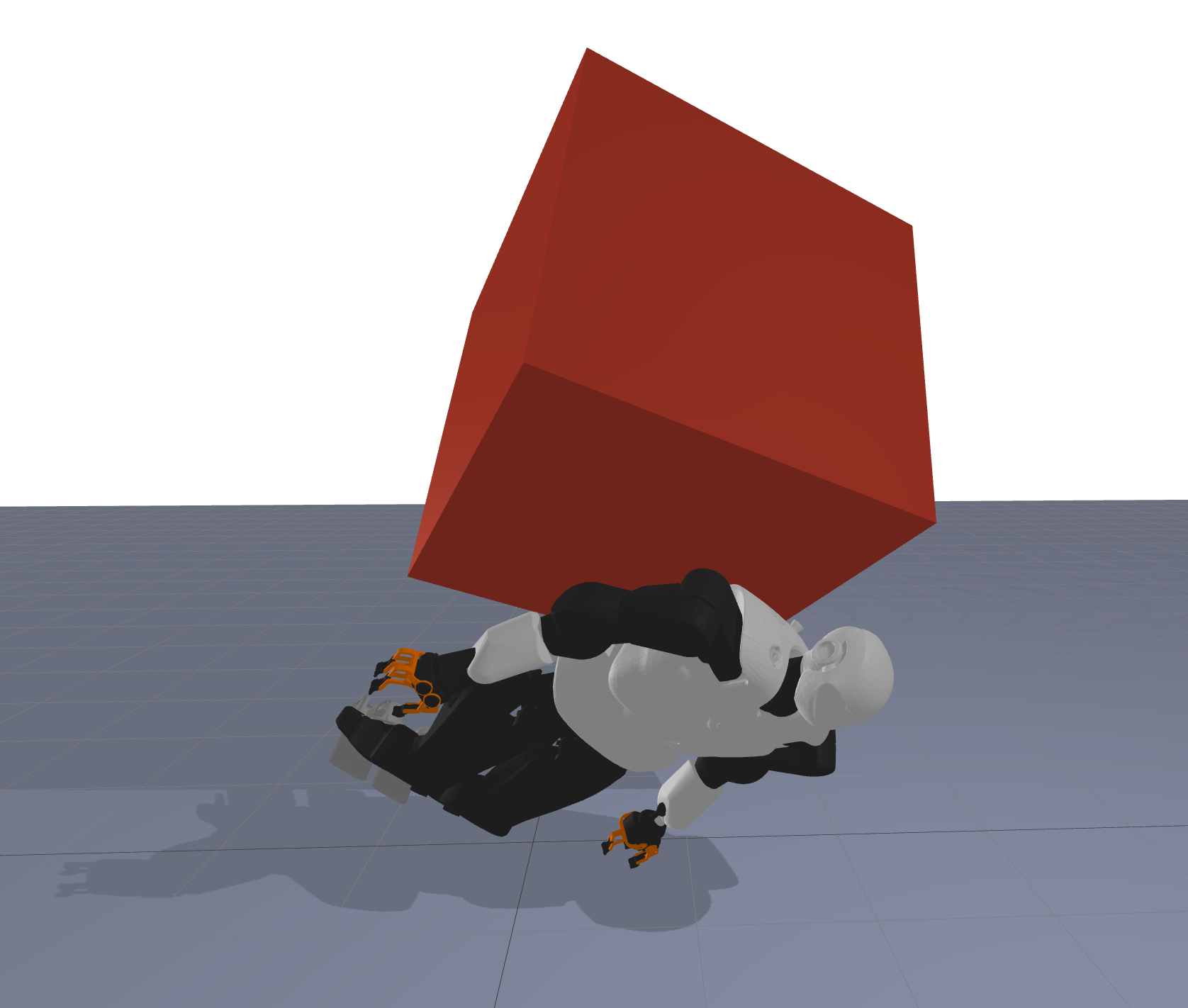}~~
    \includegraphics[width=.3\linewidth]{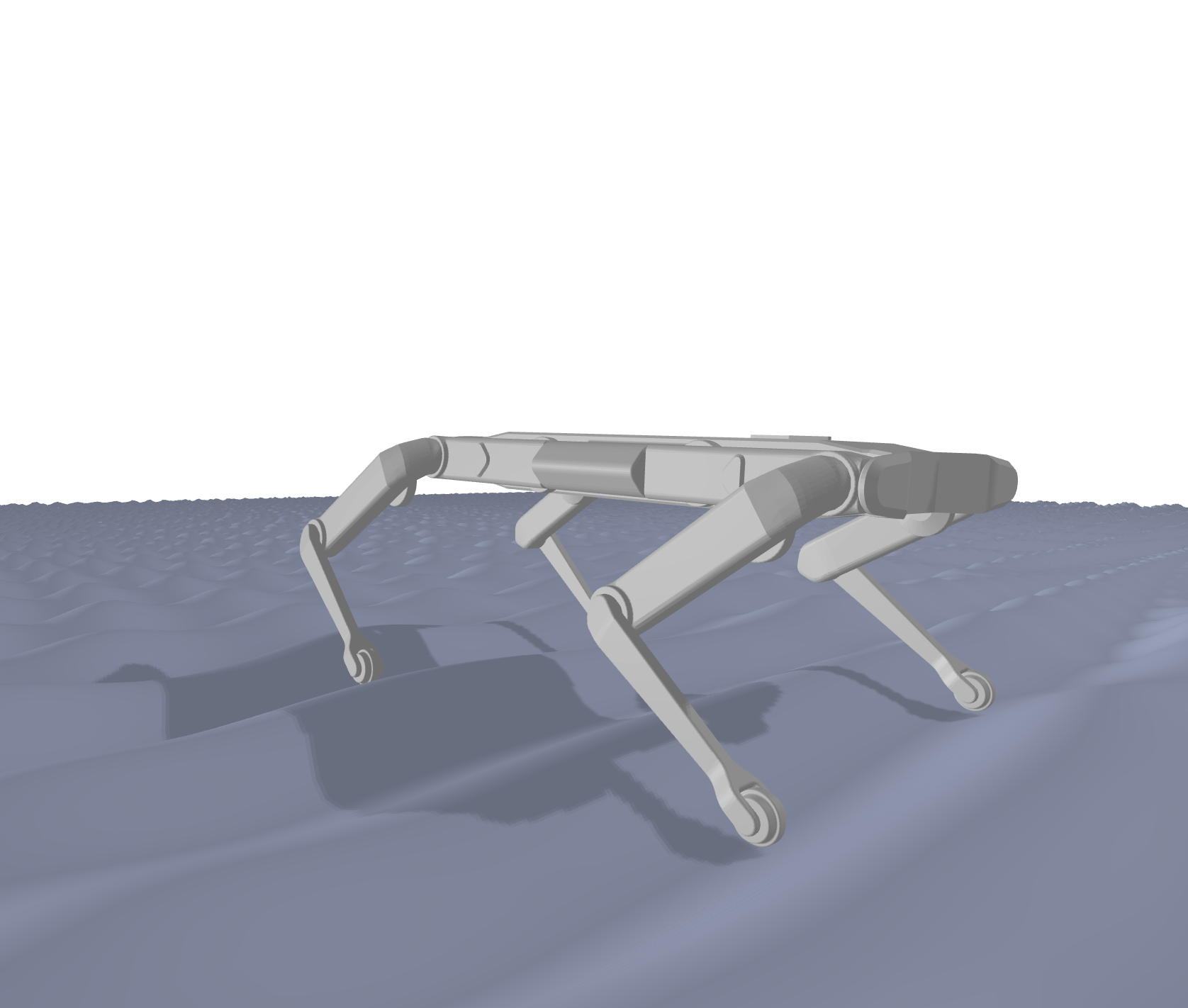}~~
    \includegraphics[width=.3\linewidth]{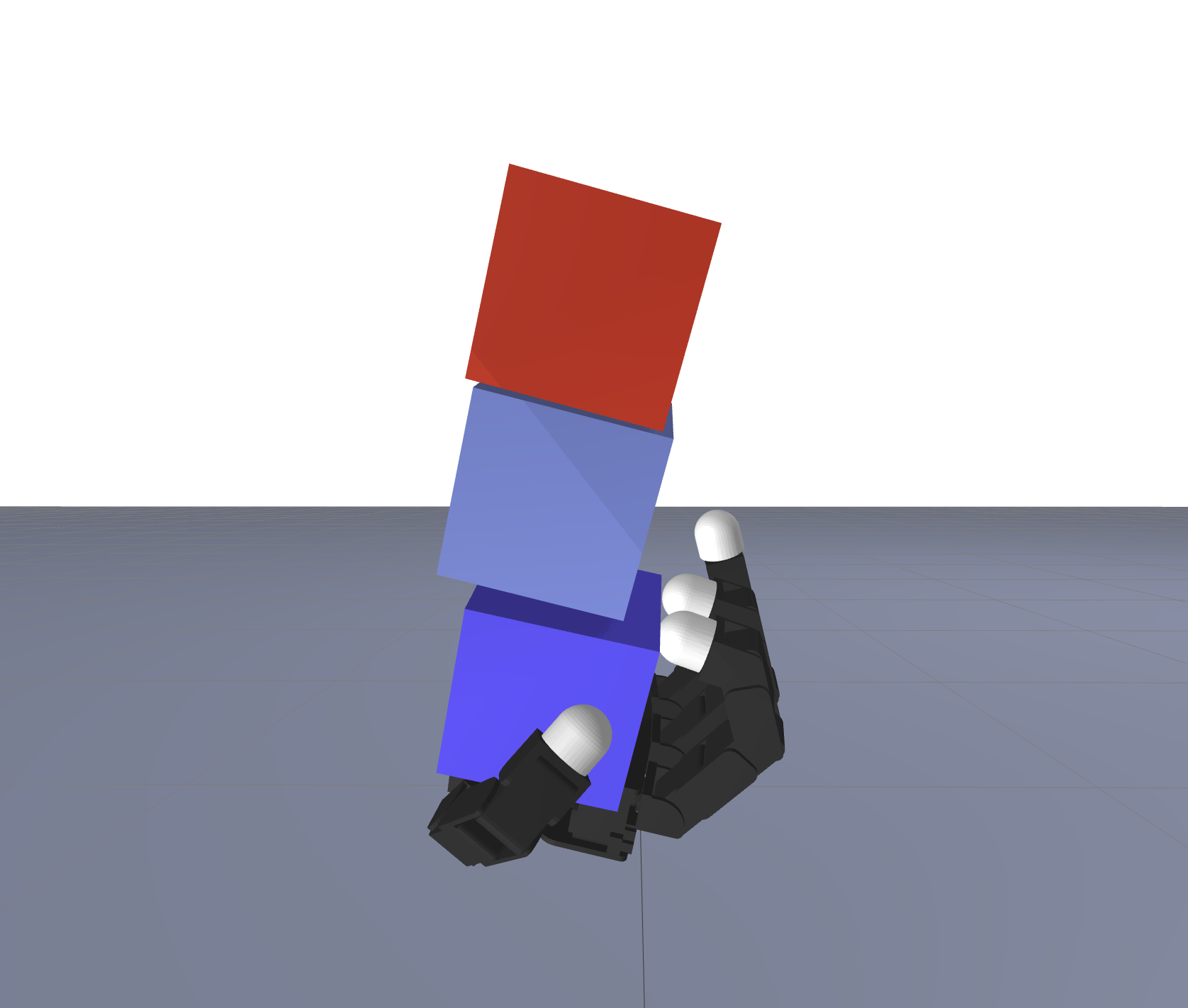}
    \caption{\textbf{Robotics systems.} We evaluate our approach on common robotics scenarios occurring during locomotion, e.g., Talos falling on the ground (\textbf{left}), Solo walking on steep terrain (\textbf{center}), and manipulation, e.g., Allegro hand interacting with cubes (\textbf{right}).
    }
    \label{fig:robotics_systems}
    \vspace{-0.0cm}
\end{figure*}

In this section, we evaluate our approach to challenging robotics scenarios and benchmarks from the computational mechanics community.
In particular, we measure performances in terms of the number of iterations required to converge and computational timings.
We also validate our inverse dynamics algorithm by performing a control task involving contact interactions with a UR5 arm.
Additional results are available in the companion video (\url{https://youtu.be/i_qg9cTx0NY?si=NGtx1tiYrIGtHXSK}).

\subsection{Simulation benchmarks: standard robotics systems}

Our solver has been implemented in C++, leveraging the Eigen library~\cite{eigenweb} for efficient linear algebra, the Pinocchio framework \cite{pinocchioweb} for efficient rigid body algorithms, which comes with the HPP-FCL library~\cite{hppfclweb} for fast collision detection.
As baselines, we have also implemented the over-relaxed PGS algorithm~\cite{jourdan1998gauss} and used the state-of-the-art SCS solver~\cite{odonoghue:21} to solve the CCP relaxation.
Our experiments are done on a MacBook Pro with a M1 Max CPU. 
For all the benchmarks, we use a time step $\Delta t$ of $1$ms and set the proximal value $\eta$ to $10^{-6}$.\\

\noindent
\textbf{Convergence analysis.} We evaluate the convergence speed of our algorithm by monitoring the evolution of the convergence criteria $\|\text{r}_k^\text{dual}\|_\infty$ and $\|\text{r}_k^\text{comp}\|_\infty$ across iterations ($\|\text{r}_k^\text{prim}\|_\infty$ being null due to projection steps).
The study is done on two different contact problems.
The first one is a stack of rigid boxes of different masses (between $1$ and $10^4$kg) hit by a ball of $10^3$kg (Fig.~\ref{fig:header}).
Stacking objects of high mass ratios induces bad conditioning of the Delassus matrix $G$ (these ratios are of $10$ for the Talos scene and $10^4$ for the wall of cubes). 
Therefore, this scene allows us to evaluate the numerical stability of the solvers.
For the second one, we study a problem obtained with the humanoid robot Talos~\cite{stasse2017talos} falling on the ground, whose 36-dof kinematic chain induces a complex inertial coupling between all the contact points through the Delassus matrix.
Additionally, we use these two scenarios to perform an ablation study on the benefits of using the spectral update rule for the ADMM parameter adaption.

\begin{figure}
    \centering
    \includegraphics[width=.48\linewidth]{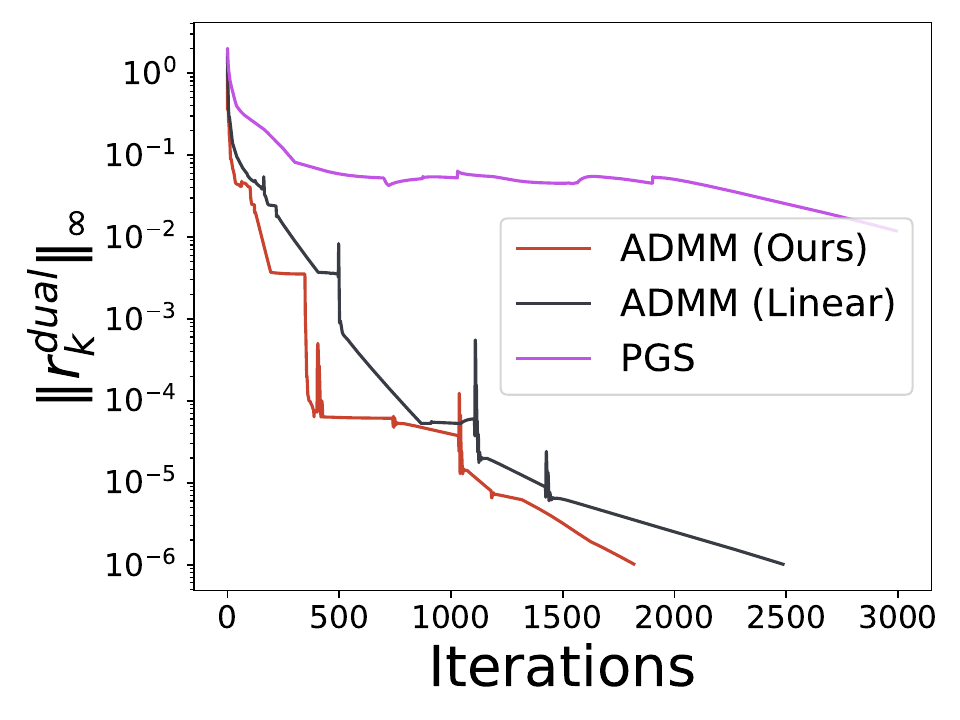}
    \includegraphics[width=.48\linewidth]{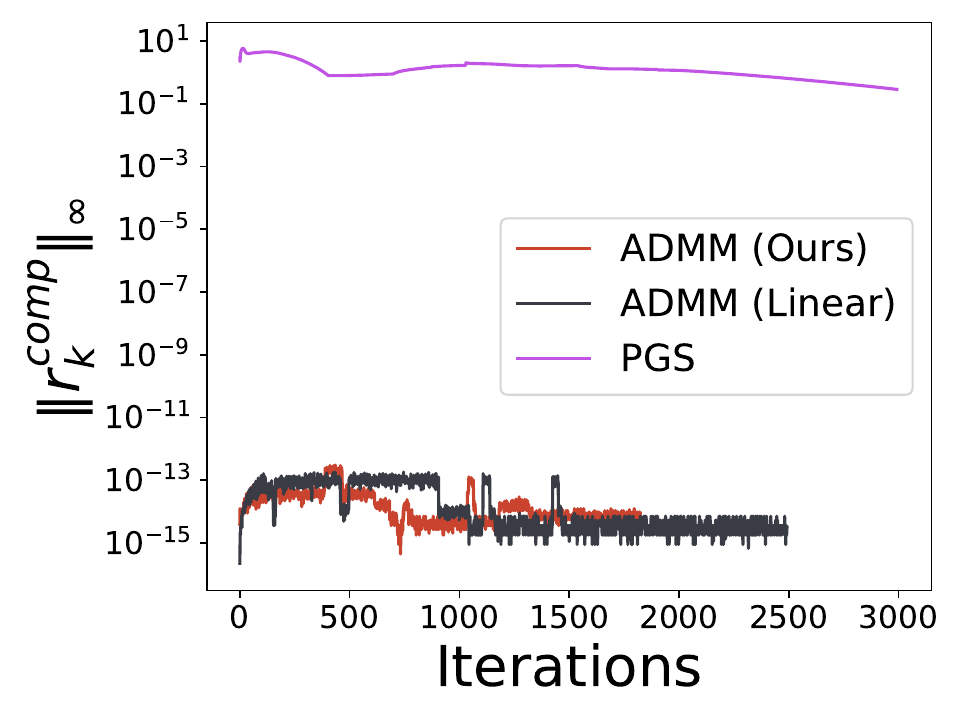}
    \includegraphics[width=.48\linewidth]{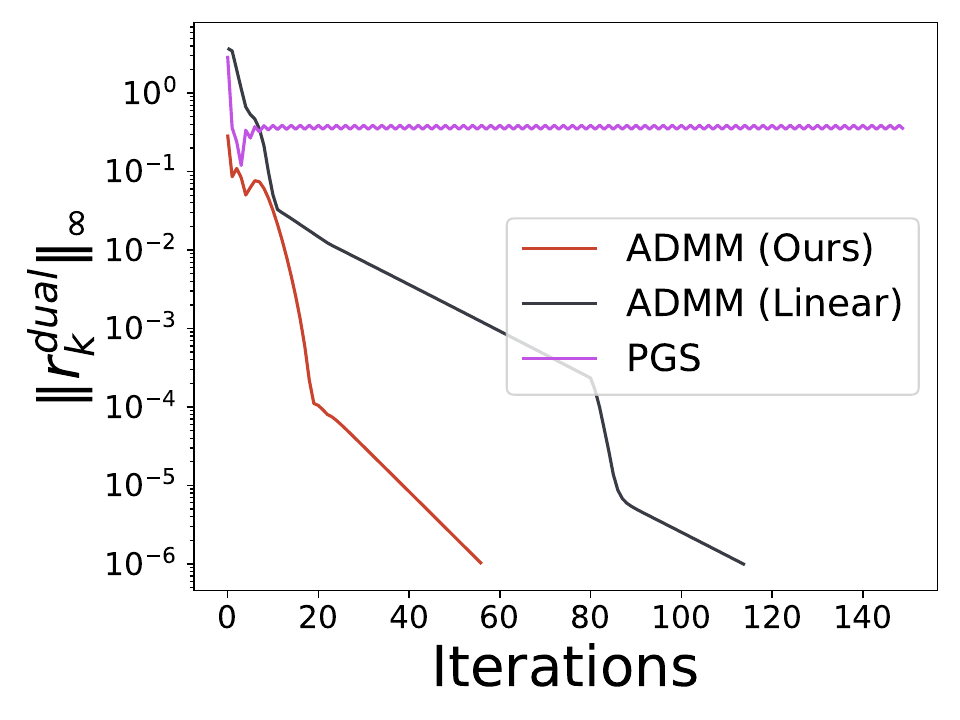}
    \includegraphics[width=.48\linewidth]{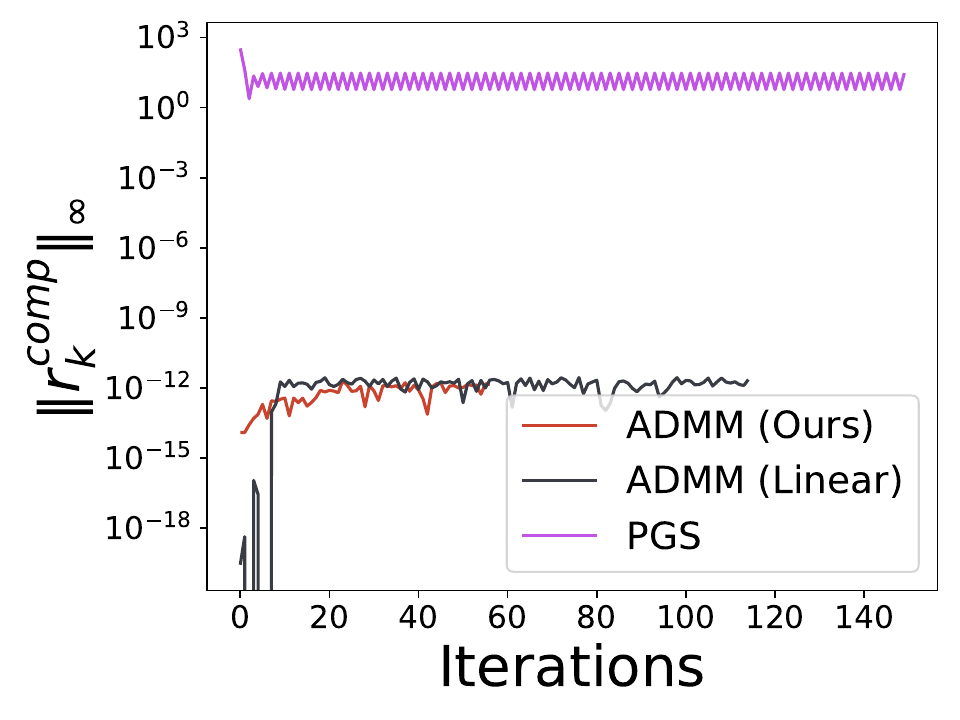}
    \caption{\textbf{Convergence analysis.}
    Convergence of the dual feasibility (\textbf{left}) and complementarity (\textbf{right}) is monitored across iterations on the contact problems obtained from simulating a stack of cubes (\textbf{top row}) and the Talos robot (\textbf{bottom row}).
    }
    \label{fig:robotics_iterations}
    \vspace{-0.0cm}
\end{figure}

As shown by Fig.~\ref{fig:robotics_iterations}, the per-contact strategy of PGS loops; this undesired behavior is explained by the fact that these problems involve high mass ratios and strongly coupled contact forces.
On the opposite, our ADMM-based algorithm exploits the Cholesky decomposition of $G_{\rho,\eta}^R$ which allows it to be insensitive to the conditioning and to capture the coupling between individual contact problems.
Figure \ref{fig:robotics_iterations} also demonstrates how leveraging the spectral information to update adapt $\rho$ leads to an improved convergence rate w.r.t. the linear update rule.
Therefore, our approach can efficiently solve the two problems at a high precision threshold $\epsilon_\text{abs} = 10^{-9}$.\\

\noindent
\textbf{Timings.} When evaluating timings, we cover a range of contact types that aim to be wide enough to represent robotics applications.
To do so, we consider four distinct scenarios (Fig.~\ref{fig:header},\ref{fig:robotics_systems}).
For the first two, we reuse the setups of the convergence analysis: a stack of boxes of different masses (around 60 contact points) and a humanoid falling on the ground (around 10 contact points).
As a third experiment, we simulate an Allegro hand holding a stack of cubes, a classical setup in manipulation applications.
Finally, we evaluate our approach on a more dynamic task: a quadruped moving on a steep terrain (Fig.~\ref{fig:robotics_systems}). 
The walking motion is generated via an MPC controller, which produces reactive behavior and results in a wide range of contact types, notably many breaking and sliding contacts.

\begin{figure}
    \centering
    \includegraphics[width=.95\linewidth]{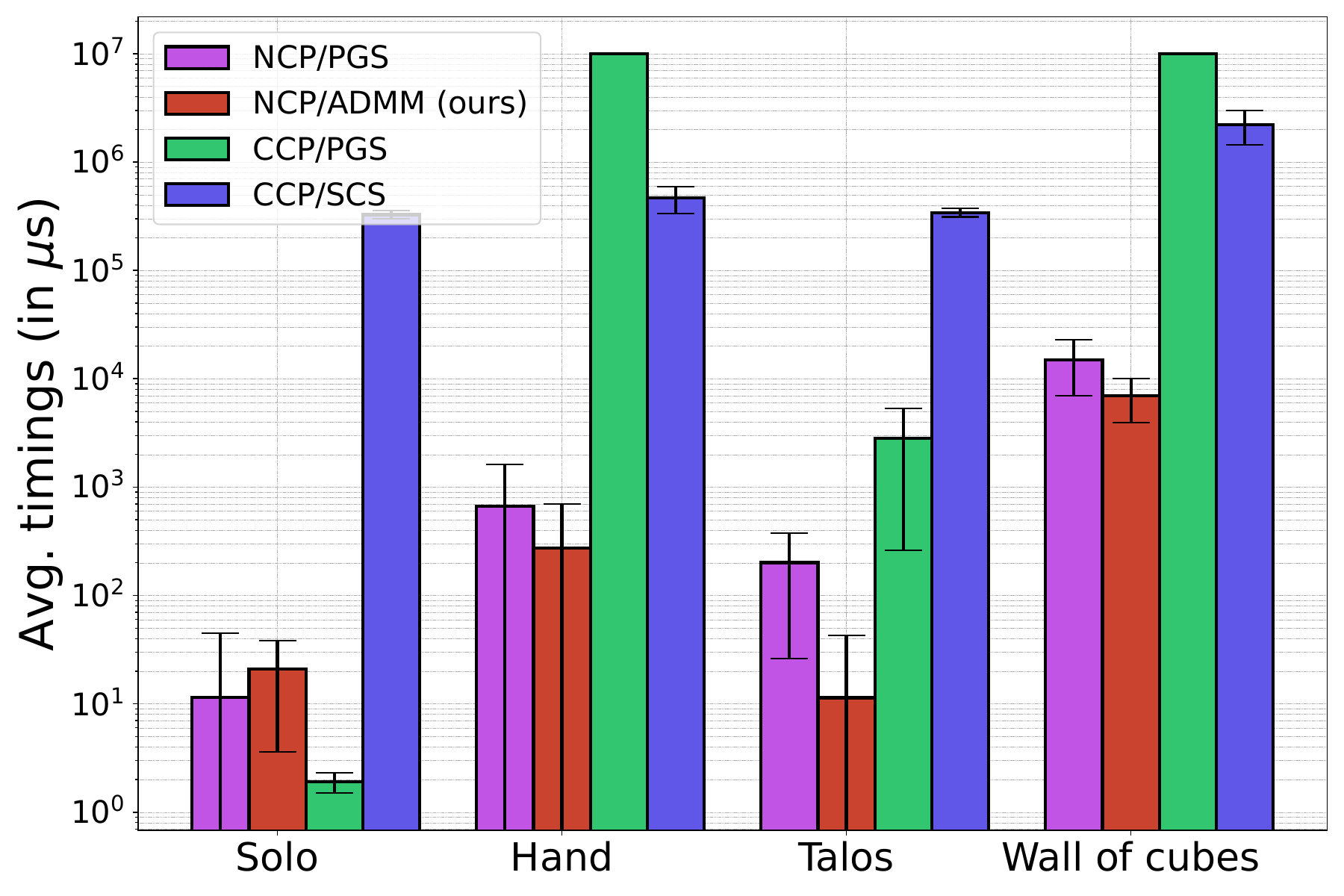}
    \caption{\textbf{Timings on robotics systems.} Thanks to its combination of automatic spectral adaptation (Sec.\ref{subsec:admm_update_parameter}) and sparse Cholesky update (Sec.~\ref{subsec:sparsity}), our algorithm (in red) yields a stable behavior across the various robot simulations.
    }
    \label{fig:robotics_timings}
    \vspace{-0.0cm}
\end{figure}

For each benchmark, we average on a trajectory the computational time necessary to reach a fixed precision ($\epsilon_\text{abs} = 10^{-4}$) or to hit the maximum number of iterations ($n_{\text{iter}} = 10^3$).
Figure \ref{fig:robotics_timings} summarizes our evaluation of timings on robotics systems.
We observe that our algorithm performs consistently well in different scenarios compared to the PGS and SCS algorithms, even when they operate on the relaxed CCP problem.
In particular, we note a significant performance gap between ADMM and PGS when the complexity of the structure of the inertia grows, as is the case for the hand, the humanoid, and the stack of cubes.
As shown during the convergence analysis, the complex coupling induced by $G$ can slow down per-contact approaches and even hinder convergence.
It is also worth noting that our algorithm does not require any hyperparameter tuning across the various considered scenarios.
Indeed, $\rho$ is automatically scaled via the spectral rule, which induces a reduced computational overhead when combined with rigid body algorithms for the Cholesky updates. \\

\subsection{Comparison against state-of-the-art physics engines}

\begin{figure}
    \centering
    \hspace{-0.1cm}
    \includegraphics[width=0.48\linewidth]{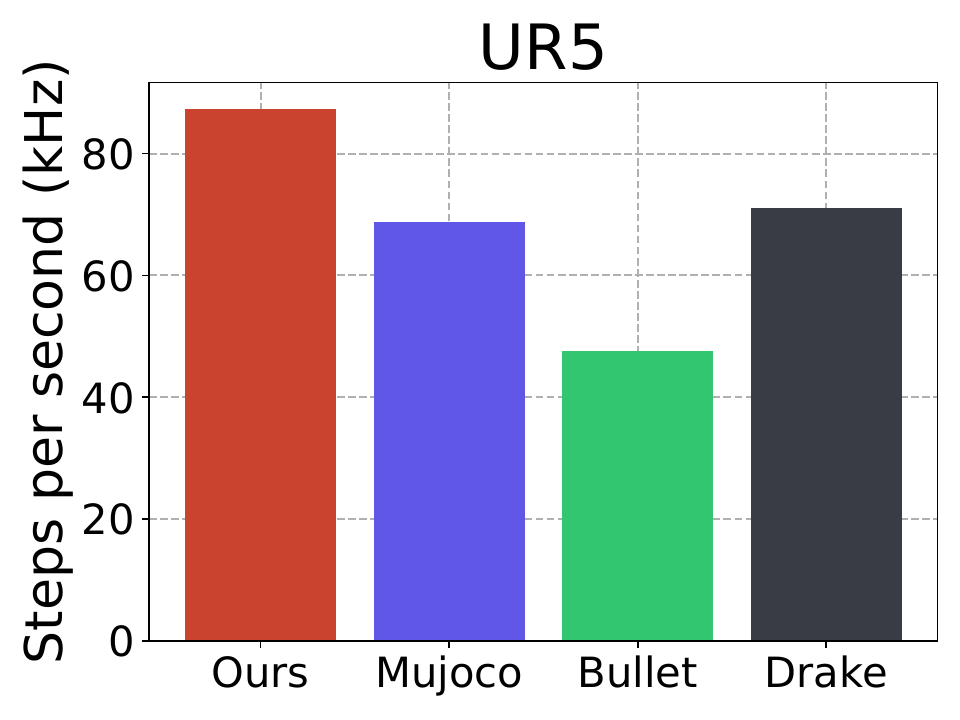}~~
    \includegraphics[width=0.48\linewidth,height=0.3\linewidth,keepaspectratio]{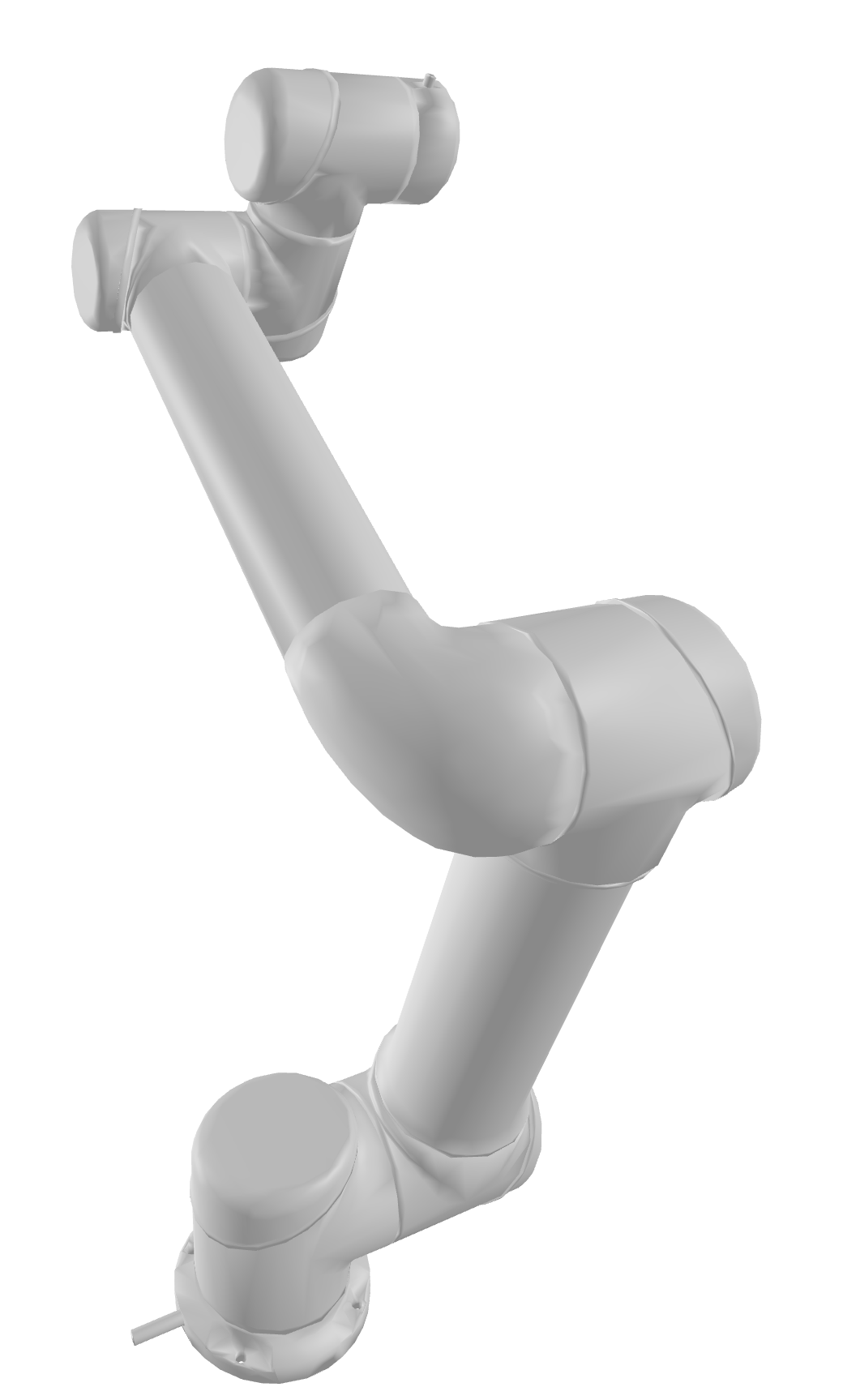}
    \\
    \hspace{-0.0cm}
    \includegraphics[width=0.48\linewidth]{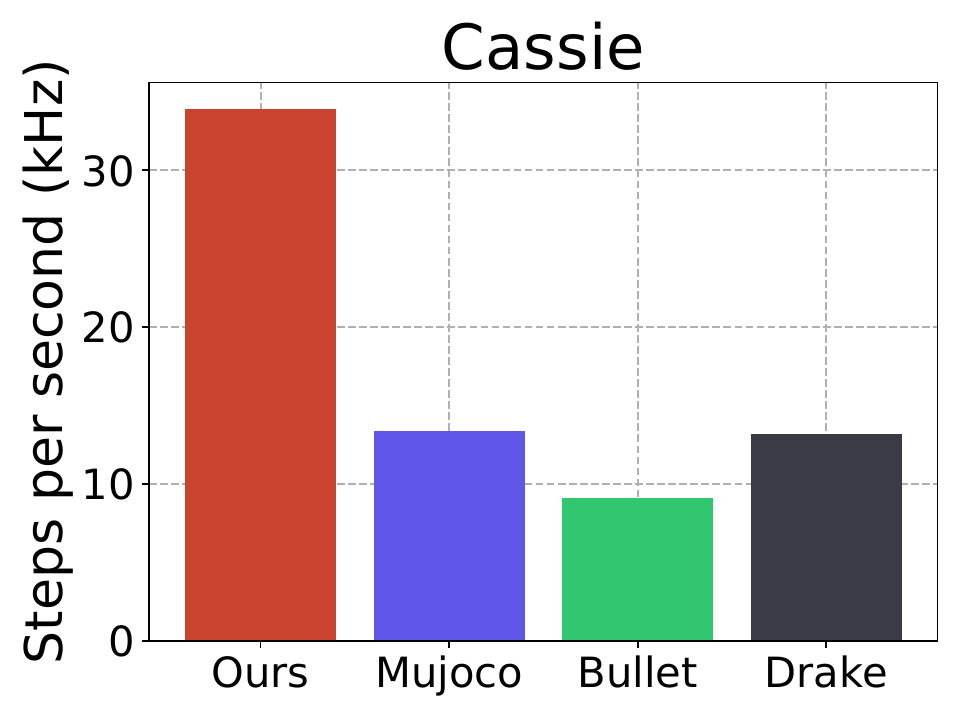}~~
    \includegraphics[width=0.48\linewidth,height=0.3\linewidth,keepaspectratio]{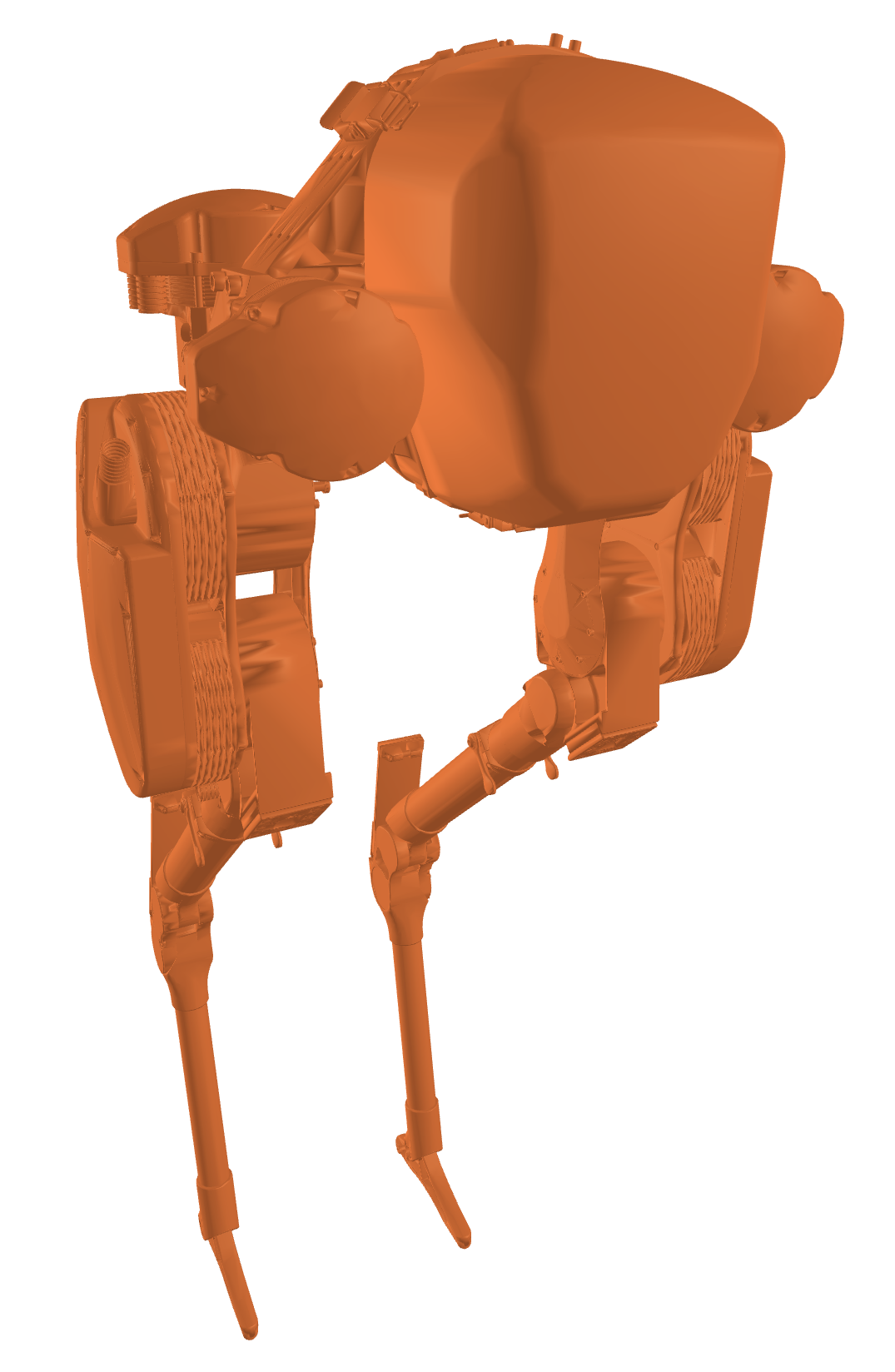}
    \hspace{-0.2cm}
    \includegraphics[width=0.48\linewidth]{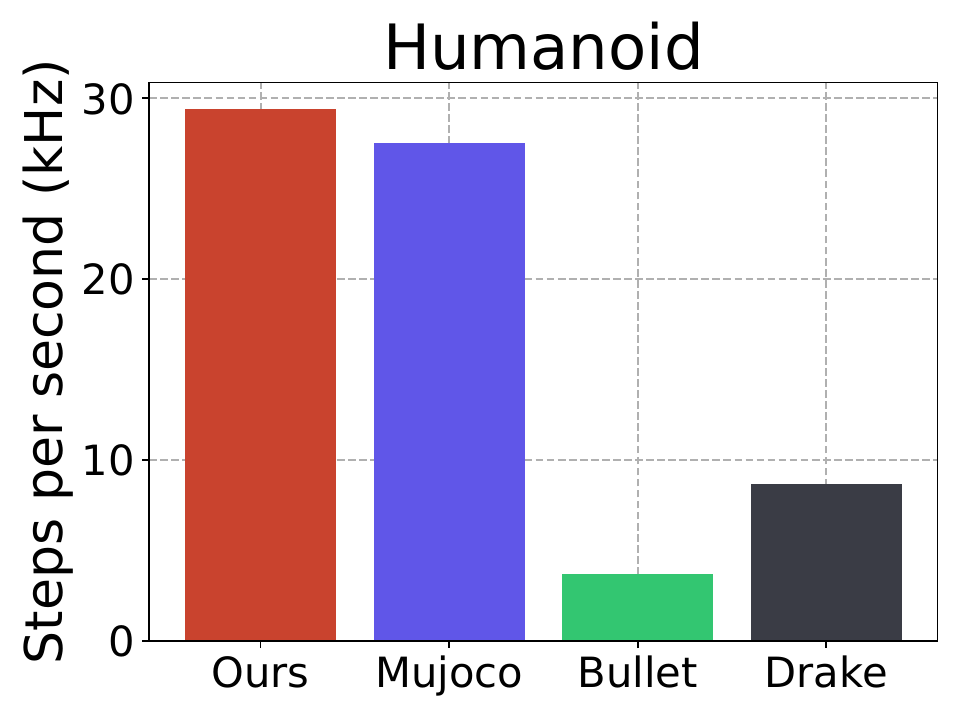}~~
    \includegraphics[width=0.48\linewidth,height=0.4\linewidth,keepaspectratio]{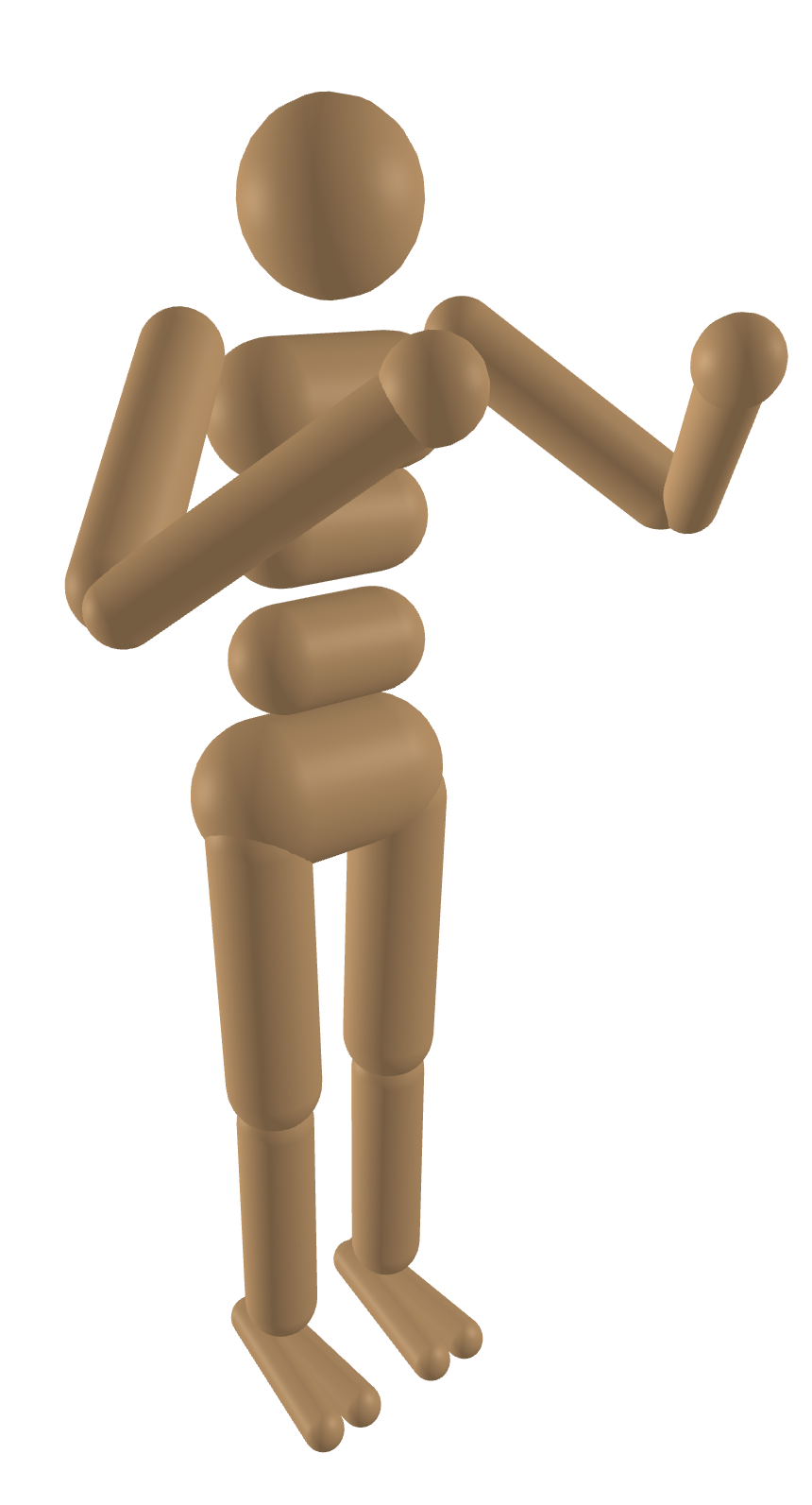}
    \\
    \caption{
    \textbf{Comparison against SOTA physics simulators.}
    Performances are reported for a UR5 arm (\textbf{top}), and a Cassie robot (\textbf{middle}), the MuJoCo simple humanoid (\textbf{bottom}).
    }
    \label{fig:simulation_benchmarks}
\end{figure}

We implemented our algorithm in a C++ simulation loop, building on top of Pinocchio~\cite{pinocchioweb} for rigid body dynamics and HPP-FCL~\cite{hppfclweb} for collision detection.
We evaluate our approach against the state-of-the-art simulators MuJoCo, Drake, and Bullet on three scenarios of increasing complexity in terms of degrees of freedom: a UR5 robotic arm, a Cassie robot, and the MuJoCo humanoid.
For each benchmark, we simulate the robot for 2 seconds without any actuation or damping in the joints. 
For all simulators, joint limits are not taken into account.
The MuJoCo Humanoid is made of capsules for collisions, while the UR5 and Cassie robots use convex meshes (which can reach up to 1000 vertices per body).
All possible pairs of bodies are considered for collision detection, except the successive pairs in the kinematic chain.
Performance is measured by the number of time steps per second on a single thread of an Apple M1 Max CPU.
The results are reported in Fig.~\ref{fig:simulation_benchmarks}.
The ADMM solver embedded in a simulation loop depicts competitive results compared to alternative solutions in the robotics literature.

It is worth noticing that this is a preliminary benchmark. 
A more detailed one (involving additional metrics, more physical systems, and comparisons to real data) would benefit the robotics community but would require a dedicated study.
We leave this benchmarking study as future work.

\subsection{Complex scenarios of computational mechanics}

In this section, we leverage the collection of complex contact problems provided in the FCLIB benchmark collections~\cite{acary2014fclib}.
These problems correspond to a set of various mechanical problems identified as challenging by the computational mechanics community.
We focus on three main categories of problems: \textit{BoxesStack}, \textit{Chain}, and \textit{Capsules}.
The size and properties of this dataset are reported in Tab.~\ref{tab:fclib}.\\

\noindent
\textbf{Timings.}
We benchmark our solver against PGS on the three aforementioned scenarios.
We request a precision of $\epsilon_\text{abs} = 10^{-6}$ on both solvers.
Due to the size of the problems, we exploit the sparse Cholesky solver coming with Eigen to account for the sparsity of the contact problems.

\begin{figure*}
    \centering
    \includegraphics[width=.32\linewidth]{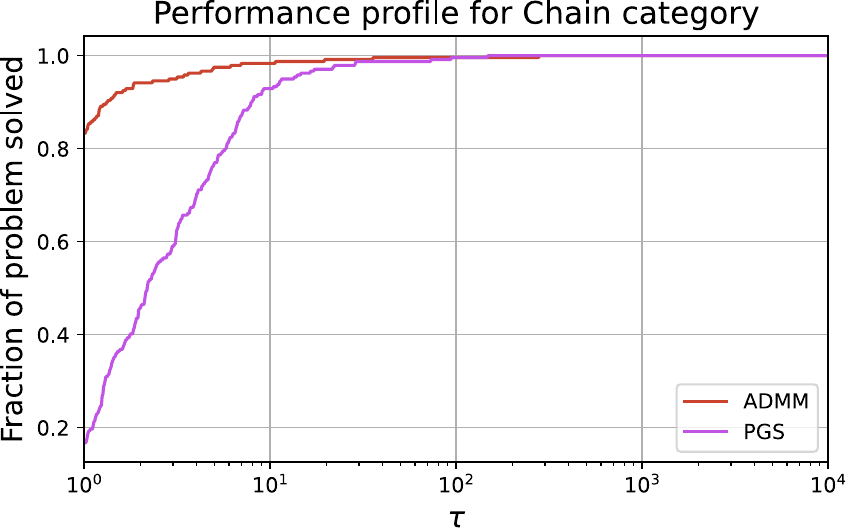}
    \includegraphics[width=.30\linewidth]{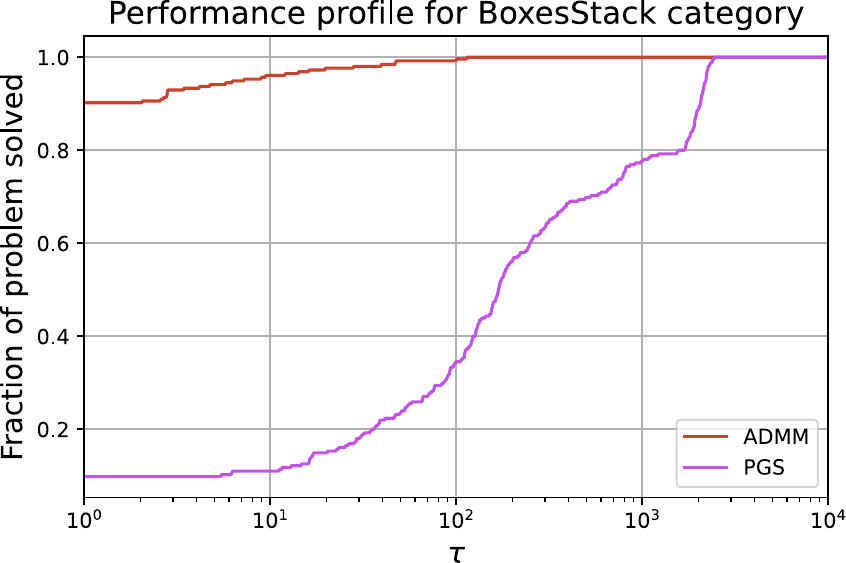}
    \includegraphics[width=.28\linewidth]{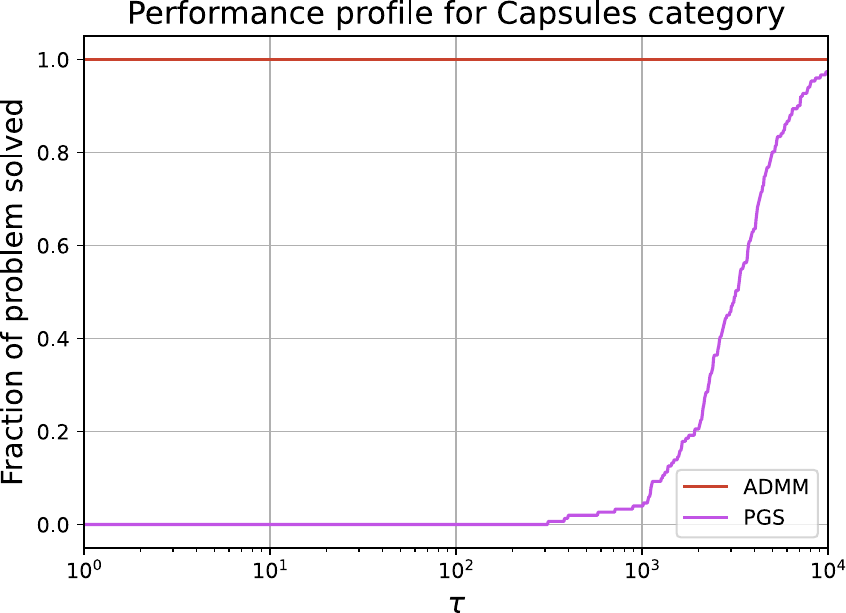}
    \caption{\textbf{Performance profiles} curves on a subset of the FCLIB dataset.
    }
    \label{fig:exp_mec}
    \vspace{-0.0cm}
\end{figure*}

We use performance-profile distributions~\cite{dolan2002benchmarking}, a metric to fairly compare optimization solvers in the optimization community. 
More precisely, for a given solver, it measures the ratio of problems solved which are $\tau$ times slower than the best solver for a given problem. 
We refer to~\cite{dolan2002benchmarking} for further details.
The performance-profile distributions are reported in Fig.~\ref{fig:exp_mec} for the three categories. 
These plots show that our solver significantly outperforms PGS on the complex contact problems collected in the FCLIB dataset.
On problems from \textit{BoxesStack} and \textit{Capsules}, we observe that PGS is not able to converge to the desired accuracy and reach the maximum number of iterations $n_\text{iter}$ set to $20.000$.\\

\begin{table}[]
    \centering
    \begin{tabular}{r|cccc}
        \toprule
         Category & \# Problems  & $n_c$ & Dofs & friction $\mu$ \\
         \midrule 
        \rowcolor{pastelblue}
         \textit{BoxesStack} &  255 & [0:200] & [6:300] & 0.7\\
         \textit{Chain} & 242 & [8:28] & [48:60] & 0.3 \\
        \rowcolor{pastelblue}
         \textit{Capsules} & 249 & [0:200] & [6:300] & 0.7 \\
         \bottomrule
    \end{tabular}
    \caption{Problems of the FCLIB dataset~\cite{acary2014fclib}.}
    \label{tab:fclib}
\end{table}

\noindent\textbf{Empirical evaluation of the spectral and linear ADMM updates.}
In section~\ref{subsec:admm_update_parameter}, we have motivated the introduction of the spectral update strategy as a way to reduce the number of Cholesky factorization updates involved in the ADMM algorithm~\ref{alg:admm_ncp}, which is the most computationally demanding part of the ADMM solver.
We evaluate this assertion by evaluating the mean number and the standard deviation of Cholesky updates over the \textit{BoxesStack}, \textit{Chain}, and \textit{Capsules} problem categories. We notably vary the linear parameter $\tau$ and spectral parameter $p$.

The results are reported in Tab.~\ref{tab:mean_cholesky_updates}.
Depending on the problem category, the linear update strategy tends to produce higher standard deviations and mean numbers of Cholesky updates than the spectral rule.
The spectral rule provides the most consistent statistics with lower updates and is less sensitive to the class of problems to solve, with an average of a dozen Cholesky updates.
We also report in Tab.~\ref{tab:mean_timings} the mean timings to solve a given problem. These timings are consistent with the mean number of Cholesky updates reported in Tab.~\ref{tab:mean_cholesky_updates}, in the sense that best timings are obtained for the spectral update rule depicting the lowest number of Cholesky updates.

\begin{table*}[t]
\centering
\caption{Comparison of the number of Cholesky updates between Linear and Spectral update strategies}
\label{tab:mean_cholesky_updates}
\begin{tabular}{r|cccc|ccc}
 {} &  \multicolumn{4}{c|}{\textbf{Linear}} & \multicolumn{3}{c}{\textbf{Spectral}}\\
\midrule
Category   & $\tau = 2$   & $\tau = 4$ & $\tau = 8$ & $\tau = 16$   & p = 0.01 & p = 0.05 & p = 0.08 \\
\midrule
\rowcolor{pastelblue}
\textit{BoxesStack}   &  
$9.20 \pm 2.48$ & $6.12 \pm 1.75$ & $5.85 \pm 2.47$ & $6.12 \pm 2.40$  & 
$11.26 \pm 4.99$ & $5.02 \pm 2.40$ & $\bm{3.97 \pm 1.76}$ \\
\textit{Chain}   &  
$5.74 \pm 1.53$ & $8.76 \pm 69.93$ & $33.7 \pm 212.77$ & $25.48 \pm 154.88$  & 
$7.87 \pm 6.55$ & $\bm{2.76 \pm 1.65}$ & $3.61 \pm 20.48$ \\
\rowcolor{pastelblue}
\textit{Capsules}   & 
$4.58 \pm 1.71$ & $3.09 \pm 2.01$ & $3.60 \pm 2.94$ & $4.6 \pm 2.06$  & 
$5.57 \pm 4.85$ & $2.30 \pm 1.50$ & $\bm{2.12 \pm 1.01}$ \\
\bottomrule
\end{tabular}
\end{table*}

\begin{table*}[t]
\centering
\caption{Comparison of the mean timings, in ms, between Linear and Spectral update strategies}
\label{tab:mean_timings}
\begin{tabular}{r|cccc|ccc}
 {} &  \multicolumn{4}{c|}{\textbf{Linear}} & \multicolumn{3}{c}{\textbf{Spectral}}\\
\midrule
Category   & $\tau = 2$   & $\tau = 4$ & $\tau = 8$ & $\tau = 16$   & p = 0.01 & p = 0.05 & p = 0.08 \\
\midrule
\rowcolor{pastelblue}
\textit{BoxesStack}   &  
$2.16 \pm 1.80$ & $1.73 \pm 1.55$ & $1.76 \pm 1.80$ & $1.95 \pm 1.87$  & 
$2.56 \pm 2.03$ & $1.55 \pm 1.60$ & $\bm{1.51 \pm 1.76}$ \\
\textit{Chain}   &  
$0.527 \pm 0.417$ & $0.637 \pm 0.373$ & $1.61 \pm 7.84$ & $1.34 \pm 6.82$  & 
$0.445 \pm 0.53$ & $\bm{0.321 \pm 0.304}$ & $0.494 \pm 0.301$ \\
\rowcolor{pastelblue}
\textit{Capsules}   & 
$2.86 \pm 1.90$ & $2.50 \pm 1.92$ & $1.95 \pm 1.35$ & $1.69 \pm 1.20$  & 
$2.17 \pm 1.38$ & $1.72 \pm 1.09$ & $\bm{1.63 \pm 1.10}$ \\
\bottomrule
\end{tabular}
\end{table*}

\subsection{Inverse dynamics}

\begin{figure}
    \centering
    \includegraphics[width=.48\linewidth]{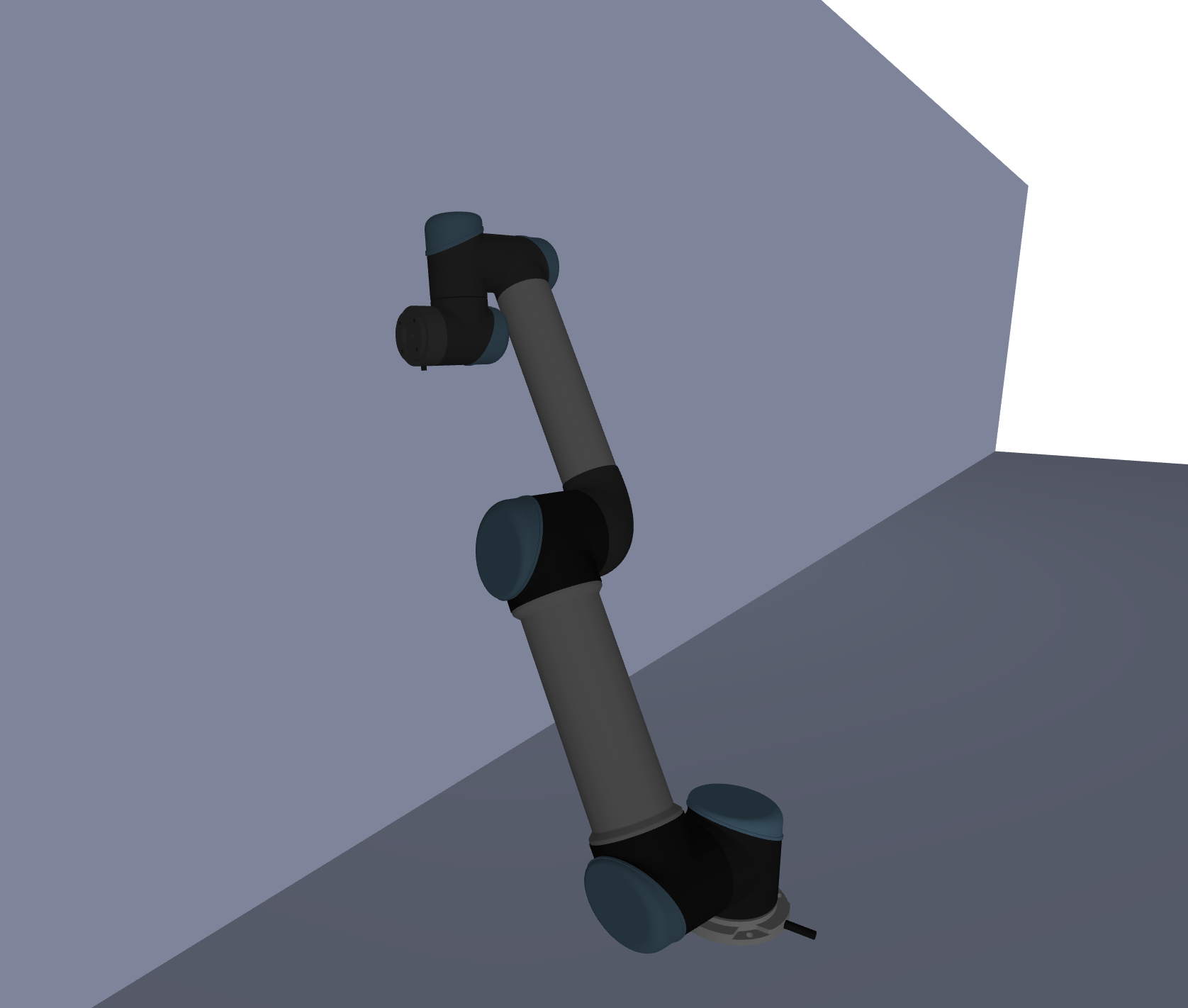}
    \includegraphics[width=.48\linewidth]{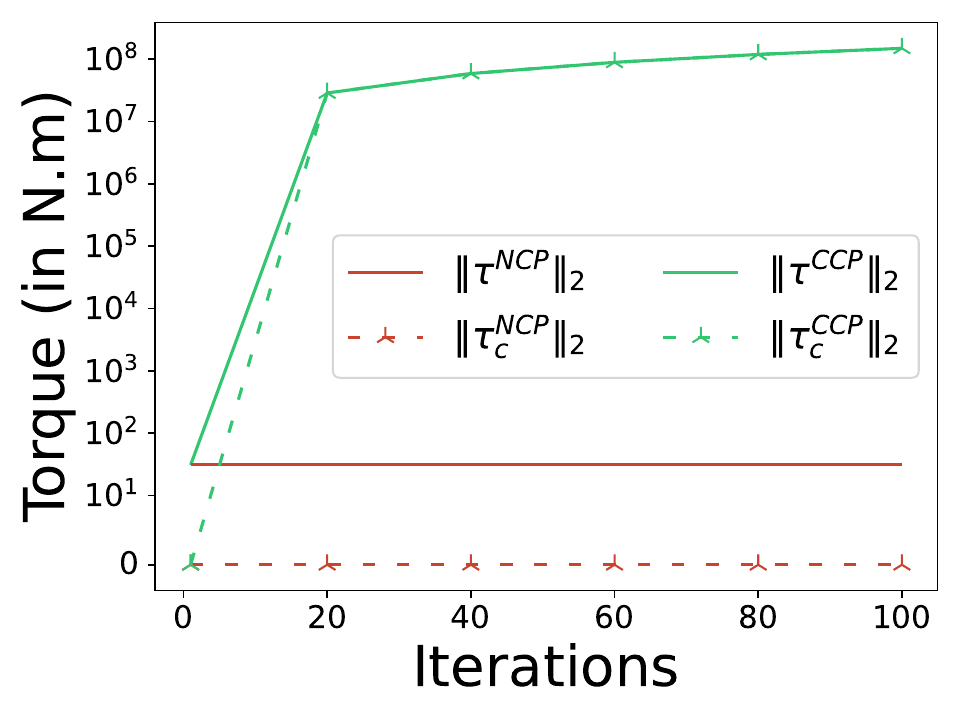}
    \includegraphics[width=.48\linewidth]{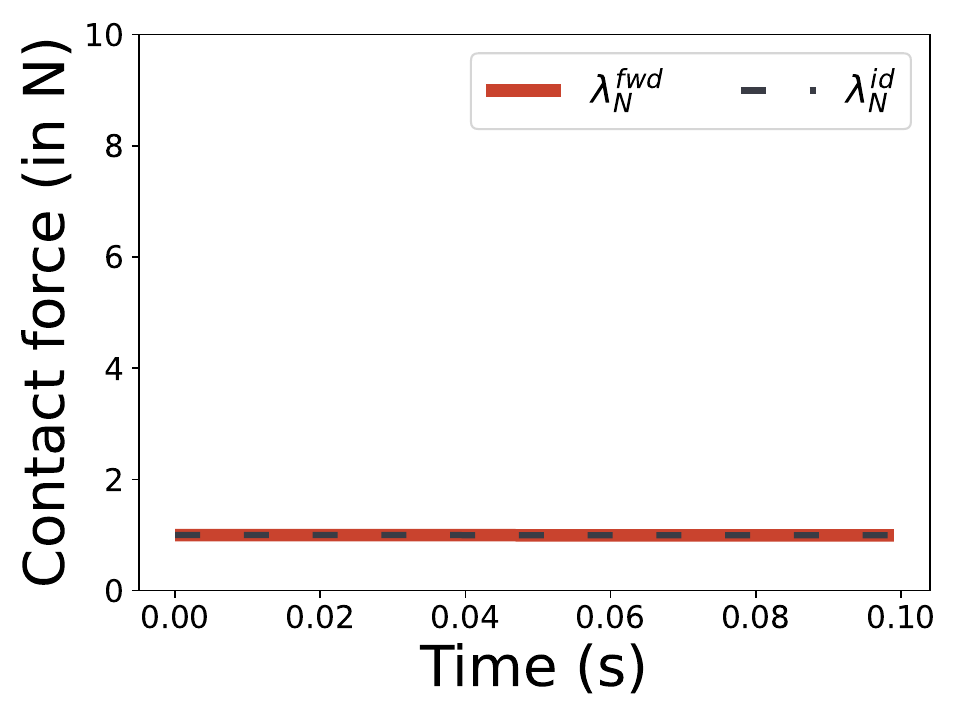}
    \includegraphics[width=.48\linewidth]{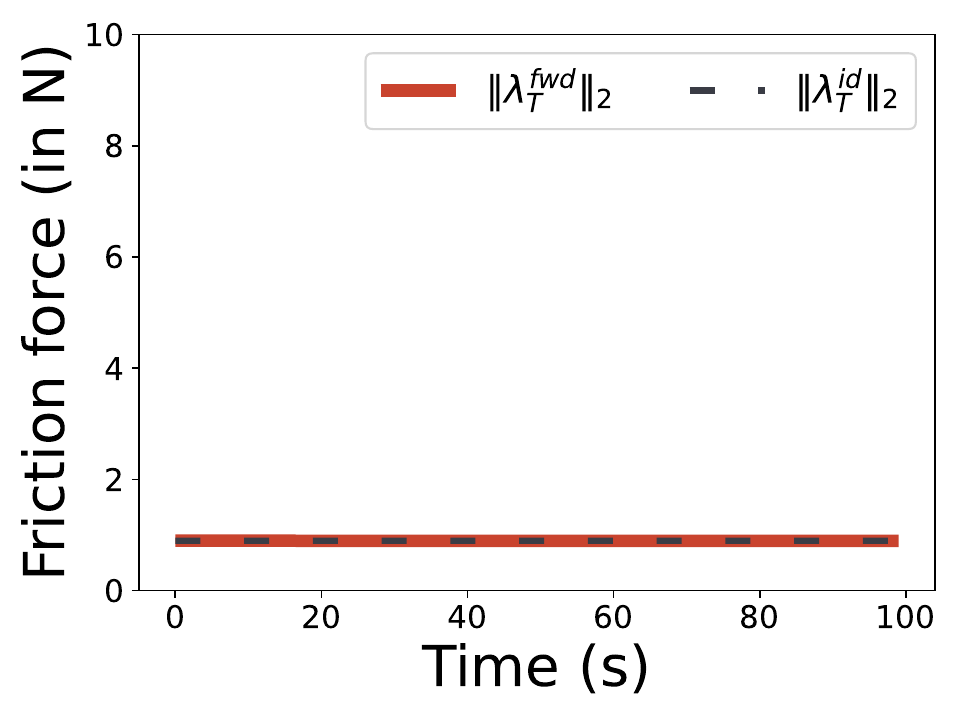}
    \caption{\textbf{Inverse dynamics} is used to slide the end-effector of a UR5 on a wall (\textbf{top left}).
    The desired tangential contact velocity is infeasible with the CCP model which causes divergence of both the joint ($\bmtau$) and contact ($\bmtau_c$) torques, while our approach robustly converges for the NCP formulation (\textbf{top right}).
    The normal (\textbf{bottom left}) and tangential (\textbf{bottom right}) contact forces computed by the inverse dynamics algorithm are consistent with the forward dynamics solution.
    }
    \label{fig:exp_id}
    \vspace{-0.0cm}
\end{figure}

We evaluate our approach for inverse dynamics on the control problem consisting of finding the torque to slide the end effector of a robotic arm (UR5) on a wall where the contact is assumed to be purely rigid $R = 0$ (Fig.~\ref{fig:exp_id}).
The reference joint velocity $\bmv_{\text{ref}}$ is such that the contact point velocity is $\bm{c}_\text{ref} = J_c \bmv_\text{ref}$ is tangent to the wall, i.e., ${\bm{c}_\text{ref}}_N = 0$.

When setting $\rho$ to $10^{-8}$, the approach proposed in Sec.~\ref{sec:inverse} requires only one iteration to find a solution with a precision of $\epsilon_\text{abs} = 10^{-6}$.
One should note that the proximal regularization is necessary due to the rigid contact hypothesis.
On Fig.~\ref{fig:exp_id}, we evaluate the benefits of incorporating the iterative De Saxc\'e correction (Alg.~\ref{alg:prox_ID}, line~\ref{line:desaxce_id}) and work directly with the NCP~\eqref{eq:NCP} model.
When setting this corrective term to $0$ which exactly corresponds to the CCP contact model, we observe that the contact torque $\bmtau_c$ and thus the actuation torque $\bmtau$ diverge.
Indeed, for a sliding motion, $\bm{c}_\text{ref} \notin \calK^*_\mu$ which causes the desired motion to be infeasible in the sense of CCP.

\section{Discussion}
\label{sec:discussion}

The closest to our work are the approaches of MuJoCo~\cite{todorov2012mujoco} and Drake~\cite{castro2022unconstrained}, which tend to solve linear systems coupling all the contact points together, by inverting a matrix containing the Delassus contact matrix.
Unlike per-contact approaches~\cite{ode:2008,coumans2021}, we exploit the Cholesky decomposition of the Delassus matrix to jointly update contact forces, which improves robustness.
As explained in Sec.~\ref{sec:admm}, such Cholesky decomposition can be efficiently computed by leveraging the sparsity pattern induced by the kinematic trees of robots composing the scene.
Moreover, we alleviate the computational footprint by reducing the number of these Cholesky re-factorizations via a spectral adaptation of the ADMM parameter.
In terms of physical realism, our improvement is twofold: we do not relax the \textit{Signorini condition}, \textit{i.e.}, our simulator does not exhibit forces at distance, and we make it possible to simulate purely rigid systems (\mbox{$R = 0$}).
Optionally, by setting De Saxc\'e corrective term to zero, we robustly solve the convex formulation of these simulators and retrieve convergence guarantees.
For future work, further experiments should be conducted to evaluate to what extent avoiding simulation artifacts related to relaxations~\cite{horak2019similarities,lelidec2023contact} tightens the reality gap of simulators \cite{fazeli2017empirical,ma2019dense} and how a reduced gap impacts downstream tasks.

Our approach also influences the formulation of the inverse dynamics.
As shown by Todorov~\cite{todorov2012mujoco}, in the case of MuJoCo and Drake, the convex relaxation and the compliance make the dynamics invertible.
Because we also encompass rigid contacts, we cannot assume the mapping between joint velocities and contact forces to be uniquely defined (e.g., hyperstatic scenarios).
Using a proximal algorithm allows handling ill-defined cases (i.e., nonunique solution) and converges toward one possible solution of the inverse problem on forces.
Moreover, via the iterative DeSaxc\'e correction, we preserve the NCP formulation.
This enables us to invert the dynamics of motions that were previously infeasible in the sense of the CCP, for instance for sliding motions.

In a parallel line of research, a recent growing effort has been made to port classical simulators such as MuJoCo and PhysX to hardware accelerators e.g. GPUs and TPUs, which resulted in MuJoCo XLA and Isaac Gym \cite{makoviychuk2021isaac}.
These architectures provide high parallelization capabilities, but they impose hard constraints on the design of contact models and algorithms.
The approach introduced in this paper focuses on exploiting the versatility and efficiency of modern CPUs to achieve physically accurate simulation at competitive rates.
However, it seems promising for future work to adapt it in order to leverage hardware accelerators.

\section{Conclusion} 
\label{sec:conclusion}

In this paper, we have introduced an ADMM-based algorithm to solve the NCP associated with the simulation of dynamics involving rigid frictional contacts.
We have evaluated our approach to challenging benchmarks from both the robotics and computational mechanics communities.
Our rich set of experiments demonstrates that we can robustly simulate a wide range of scenarios while keeping a limited computational burden and avoiding physical relaxation.

The current approach could still be improved by gaining timings on the collision detection routine corresponding to the bottleneck.
Similarly, rigid-body dynamics algorithms for constrained dynamical systems still represent an active area of research whose improvements would directly affect contact solvers and, thus, physics simulation in robotics.
Although we did not observe cases causing our algorithm to diverge, working towards theoretical convergence guarantees could be an interesting research direction.
Moreover, our algorithm for inverse dynamics could be generalized to account for the underactuation and the unfeasible reference accelerations it can induce.

Finally, we believe these promising results are a further step towards more computationally efficient and physically consistent simulators which, due to their centrality, could positively impact the overall robotics community and related fields where efficient and reliable simulation matters.

\section*{Acknowledgements}

This work was supported in part by the French government under the management of Agence Nationale de la Recherche (ANR) as part of the ”Investissements d'avenir” program, references ANR-19-P3IA-0001 (\textsc{PRAIRIE} 3IA Institute) and ANR-22-CE33-0007 (INEXACT), the European project AGIMUS (Grant 101070165), the Louis Vuitton ENS Chair on Artificial Intelligence and the Casino ENS Chair on Algorithmic and Machine Learning.
Any opinions, findings, conclusions, or recommendations expressed in this material are those of the authors and do not necessarily reflect the views of the funding agencies.

\balance{}
\bibliographystyle{plainnat}
\bibliography{references}
\end{document}